\documentclass{article} 
\usepackage{iclr2027_conference,times}

\renewcommand{\headrulewidth}{0pt}

\usepackage{amsmath,amsfonts,bm}

\def\eqref#1{equation~\ref{#1}}

\def\1{\bm{1}}

\def\va{{\bm{a}}}

\DeclareMathAlphabet{\mathsfit}{\encodingdefault}{\sfdefault}{m}{sl}
\SetMathAlphabet{\mathsfit}{bold}{\encodingdefault}{\sfdefault}{bx}{n}

\usepackage{hyperref}
\usepackage{url}
\usepackage{booktabs}
\usepackage{graphicx}
\usepackage{wrapfig}
\usepackage{pifont}
\usepackage{multirow}
\usepackage[table]{xcolor}
\usepackage{caption}   
\definecolor{prismblue}{RGB}{235, 245, 252}

\newcommand{\cmark}{\ding{51}}
\newcommand{\xmark}{\ding{55}}

\newcommand{\gainhspace}{0.10em}
\newcommand{\gainraise}{-0.15ex}

\newcommand{\gain}[1]{%
  \ifdim#1pt<0pt
    \textcolor{red}{%
      \hspace{\gainhspace}\raisebox{\gainraise}{\scriptsize(#1)}%
    }%
  \else
    \textcolor{green!50!black}{%
      \hspace{\gainhspace}\raisebox{\gainraise}{\scriptsize(#1)}%
    }%
  \fi
}

\newcommand{\cellp}[2]{\makebox[4.8em][l]{#1#2}}   
\newcommand{\cellc}[2]{\makebox[3.8em][l]{#1#2}}   
\newcommand{\cells}[2]{\makebox[3.0em][l]{#1#2}}   

\newcommand{\wmodel}{3.0em}   
\newcommand{\wprism}{2.0em}   
\newcommand{\wcellg}{3.6em}   
\newcommand{\wcelld}{3.6em}   

\newcommand{\cellm}[1]{\parbox[c]{\wmodel}{\centering #1}}
\newcommand{\cellr}[1]{\makebox[\wprism][c]{#1}}          
\newcommand{\cellg}[2]{\makebox[\wcellg][l]{#1#2}}        
\newcommand{\celld}[2]{\makebox[\wcelld][l]{#1#2}}        

\newcommand{\wablmme}{4.8em}    
\newcommand{\wabldec}{4.6em}    
\newcommand{\wabldpg}{4.6em}    
\newcommand{\wablge}{5.0em}     

\newcommand{\ablmme}[2]{\makebox[\wablmme][l]{#1#2}}
\newcommand{\abldec}[2]{\makebox[\wabldec][l]{#1#2}}
\newcommand{\abldpg}[2]{\makebox[\wabldpg][l]{#1#2}}
\newcommand{\ablge}[2]{\makebox[\wablge][l]{#1#2}}

\title{Prism: Bidirectional Self-Teaching for Unified Multimodal Models}

\author{%
\makebox[\textwidth][c]{%
\begin{tabular}{c}
Peng Zhang$^{1,2}$,
Qian Chen$^{2}$,
Xianyin Zhang$^{2}$,
Yimin Tian$^{2}$,
Ruite Fei$^{1}$, \\
Dongyi Lv$^{1}$,
Lifan Guo$^{2\dagger}$,
Chi Zhang$^{2}$,
and Shuai Liu$^{1*}$ \\[0.8em]
$^{1}$Xi'an Jiaotong University \qquad
$^{2}$Qwen DianJin Team, Alibaba Cloud Computing
\end{tabular}%
}%
}

\begin{document}

\maketitle
\begingroup
\renewcommand{\thefootnote}{}
\footnotetext{%
  $^{*}$Corresponding author. \quad
  $^{\dagger}$Project lead.%
}
\endgroup

\begin{abstract}
Unified Multimodal Models (UMMs) aim to support visual understanding and generation within a single model. Existing post-training methods either use understanding as a fixed teacher to improve generation, or rely on ground-truth trajectories to supervise both understanding and generation. Prism instead derives bidirectional supervision from self-generated interleaved rollouts, constructing training targets according to the reliability of their text and image chains. We propose Prism, a bidirectional self-teaching post-training method for UMMs. Like a prism decomposing mixed light, Prism decouples a coupled interleaved trajectory into a text chain and an image chain, assesses the reliability of each chain, and constructs a corresponding training target from their reliability states. Experiments on both SenseNova-U1 and BAGEL show that Prism improves both understanding and generation. On the generation side, GenEval improves from 85.8 to 89.2 (+3.4), and DPG improves from 86.1 to 87.3 (+1.2) on SenseNova-U1. On the understanding side, the total MME score improves from 2282.6 to 2311.6, and MMVP improves from 58.0 to 62.0. Consistent improvements are observed on BAGEL. These results highlight that Prism enables coordinated improvements in generation and understanding by constructing reliability-guided self-training targets from self-generated trajectories, without relying on an external verifier or externally annotated corrected trajectories during target construction. Code is available at ~\url{https://github.com/aliyun/qwen-dianjin/tree/master/DianJin-Prism}.
\end{abstract}

\begin{figure}[!h]
    \centering
    \includegraphics[width=\linewidth]{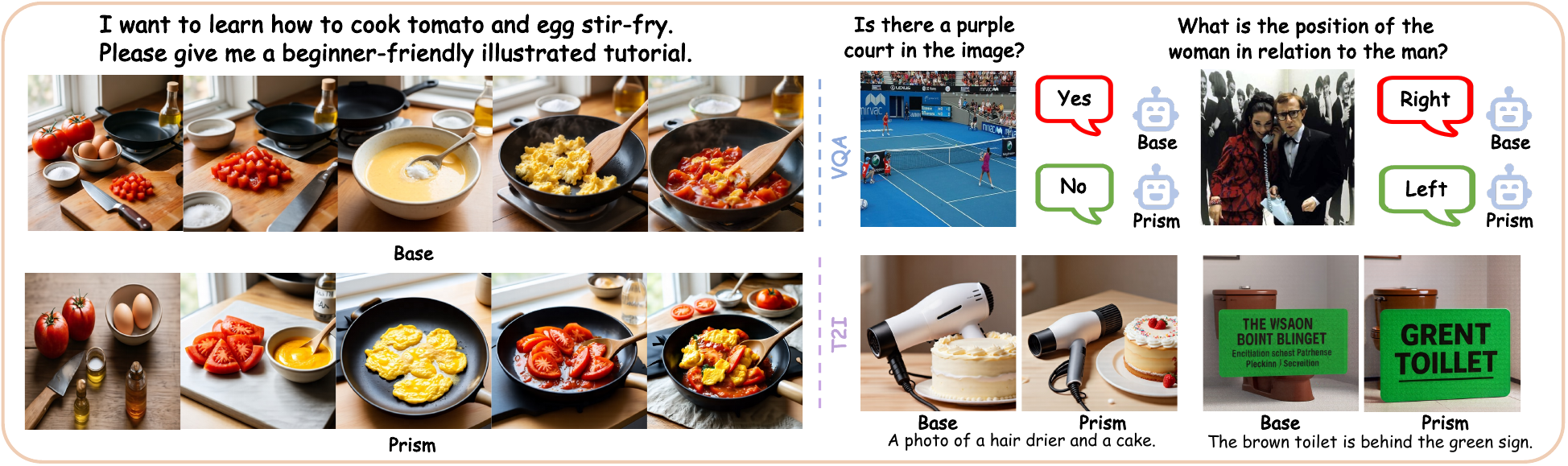}
    \caption{Capabilities of Prism. Prism improves both understanding and generation through self-teaching on interleaved text-image trajectories. The base model shows redundant steps on step 2 in the cooking tutorial, while Prism produces a coherent and correct tutorial. It also improves performance on standalone understanding and generation tasks.}
    \label{fig:motivation}
\end{figure}

\section{Introduction}
With the rapid development of large language models~\citep{gpt3,ernie,qwen} and multimodal generation models~\citep{llava,internvl2.5, qwenvl}, researchers have begun to explore unifying understanding and generation within a single model. Unlike treating understanding and generation as two separate models, Unified Multimodal Models (UMMs) integrate text, images, and their interactions into a unified architecture, enabling visual question answering and image generation to share knowledge and training signals. Recent UMMs, from Chameleon~\citep{chameleon} to SenseNova-U1~\citep{sensenova}, have demonstrated the feasibility of supporting both understanding and generation within a single model through different architectural designs. However, these models also reveal a fundamental problem: the understanding and generation capabilities within a UMM are asymmetric.

\begin{figure}[!t]
    \centering
    \includegraphics[width=\linewidth]{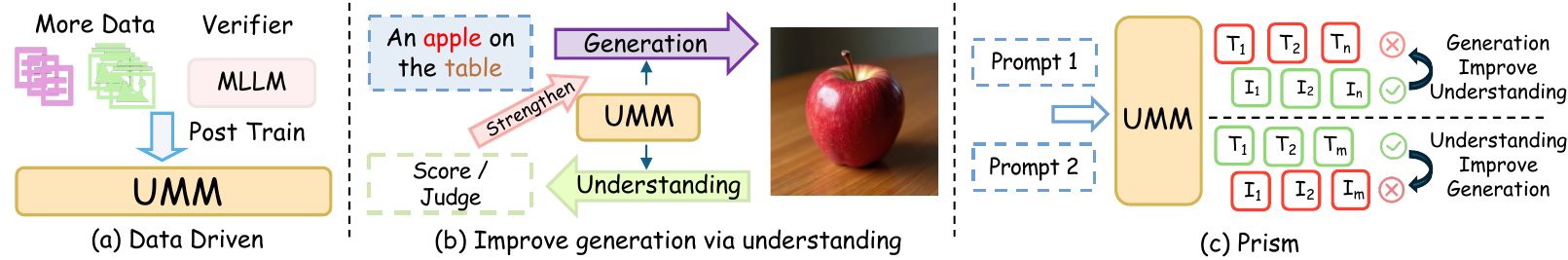}
    \caption{Difference between Prism and previous post-training methods. (a) Some prior methods are data-driven or rely on external supervision or verifiers. (b) Others improve generation through understanding in a fixed direction. (c) Prism instead enables bidirectional improvement between understanding and generation without external supervision or verifiers.}
    \label{fig:comparison}
\end{figure}

\begin{wrapfigure}{r}{0.43\columnwidth}
    \centering
    \vspace{-6pt}
    \includegraphics[width=\linewidth]{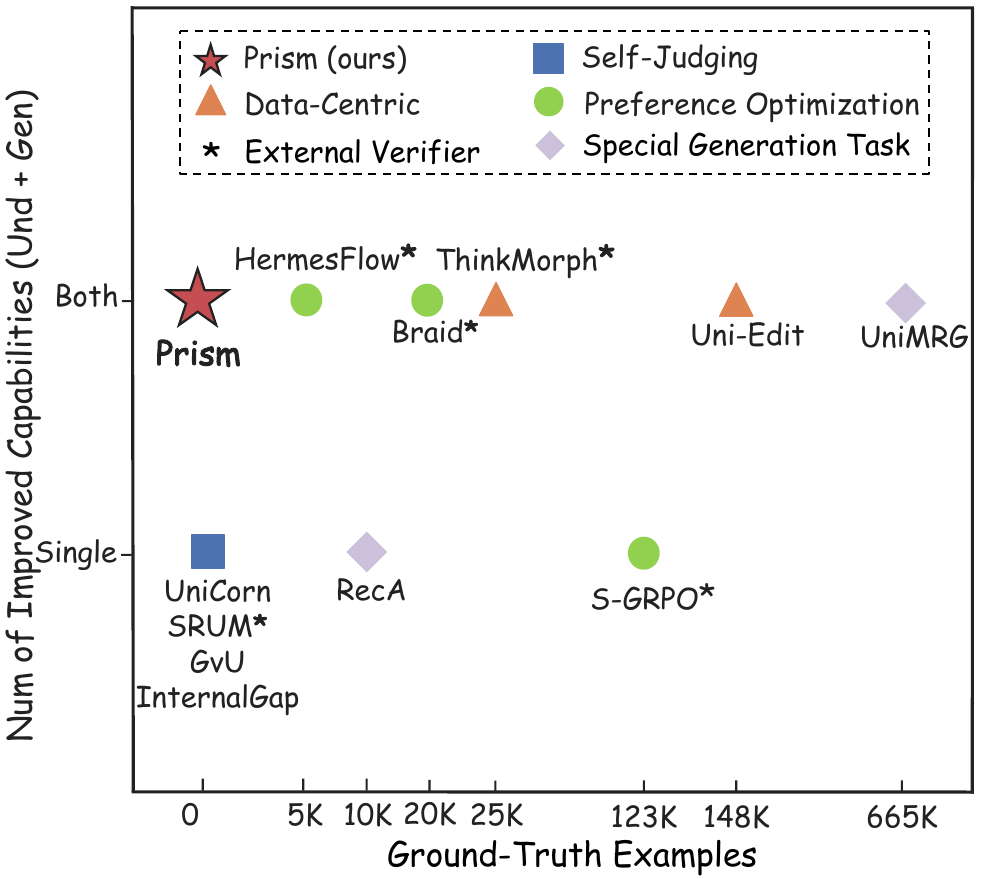}
    \caption{Prism's position by ground-truth supervision and improved capabilities.}
    \label{fig:positioning}
    \vspace{-12pt}
\end{wrapfigure}

Recent studies reveal an asymmetry in UMMs: a model can judge whether an image matches a text, yet fail to generate an image that follows the same text~\citep{reca}. Most methods exploit this in a fixed direction, using understanding to improve generation. GvU~\citep{gvu} and SRUM~\citep{SRUM} turn alignment judgments into generation rewards, Internal Gap Self-Improvement~\citep{internalgap} turns them into supervision or DPO preference pairs, and RecA~\citep{reca} reconstructs images from visual understanding embeddings. Others work in reverse, generating depth, segmentation, or reconstruction targets to sharpen perception~\citep{unimrg}, or fine-tuning on synthesized editing and interleaved reasoning traces~\citep{uniedit,thinkmorph}. A further set applies preference or policy optimization~\citep{hermesflow,unifiedgrpo,braid,sgrpo}, and UniCorn~\citep{UniCorn} explores self-play. These methods either are limited to improving generation or rely on external ground-truth pairs, reward models, or verifiers. Prism instead derives bidirectional supervision from self-generated interleaved trajectories, building training targets from text and image reliability without ground-truth trajectories or an external verifier, as illustrated in Figure~\ref{fig:comparison} and Figure~\ref{fig:positioning}.

Prism improves both understanding and generation through self-teaching on interleaved text-image tasks. In these trajectories, text and image alternate across multiple steps rather than forming a single input-output pair. Because the text and image chains are coordinated expressions of the same underlying task, each chain can be audited independently. An interleaved trajectory can therefore be decomposed into a text chain that represents understanding and an image chain that represents generation, each of which may be reliable or unreliable in different ways.

As shown in Figure~\ref{fig:motivation}, when the base model generates an interleaved cooking tutorial, the first image already shows chopped tomatoes during the ingredient-preparation stage, while the next step again depicts cutting tomatoes. For tutorial-style tasks, this is either redundant or indicates leakage of a future state into an earlier step. Conversely, although not shown in Figure~\ref{fig:motivation}, the opposite case can also occur: the image chain may still complete the task correctly even when the text chain contains erroneous reasoning or descriptions, because the image does not strictly follow the faulty text.

Therefore, the key challenge is to identify, for each sample, whether the understanding or generation side is reliable enough to teach the other. Prism addresses this challenge by performing sample-wise self-teaching over interleaved trajectories, enabling bidirectional improvement in both understanding and generation.

We propose Prism, a bidirectional self-teaching post-training method for interleaved text-image trajectories. Prism treats each self-sampled trajectory as a coupled multimodal signal and decomposes it into two independently assessable views: a text chain and an image chain. By comparing the reliability of these two chains, Prism avoids assuming a fixed supervision direction and instead determines on a per-sample basis whether text should supervise image, image should supervise text, or unreliable patterns in both chains should be suppressed.

\noindent The key contributions can be summarized as follows:
\begin{itemize}
    \item \textbf{Method:} We introduce a bidirectional post-training paradigm that selects the teaching direction for each trajectory, allowing multimodal understanding and generation to improve each other rather than imposing a fixed supervision direction.
    \item \textbf{Framework:} Prism provides a reliability-guided self-teaching framework. It constructs training targets from self-generated interleaved rollouts by assessing the reliability of their text and image chains, without ground-truth supervision or external verifiers.
\item \textbf{Performance:} Prism improves multimodal understanding and generation across different UMMs, including interleaved generation. The results demonstrate its effectiveness and generality.

\end{itemize}

\section{Related Work}

\subsection{Unified Understanding and Generation Models}

UMM architectures can be broadly categorized into four types. Bridge architectures~\citep{blip3o,openuni,joyai} retain separate understanding and generation backbones connected through lightweight interfaces. Unified Transformer architectures~\citep{chameleon,transfusion,showo,emu3,janus,janus-pro,showo2} process both modalities within a shared Transformer backbone, with variants using discrete image tokens, mixed autoregressive-diffusion objectives, or decoupled visual encoders. Mixture-of-Transformer-Experts (MoT) architectures~\citep{bagel,sensenova} allocate separate expert parameters for understanding and generation while sharing attention. Diffusion-based architectures~\citep{llada2uni} employ diffusion language models to jointly model text and images.

\subsection{UMM Post-Training}
Existing UMM post-training methods follow four main directions. Preference optimization and reinforcement learning use externally scored pairs in HermesFlow~\citep{hermesflow}, reward components in Unified Interleaved GRPO~\citep{unifiedgrpo} and BRAID~\citep{braid}, or ground-truth trajectories for VLM adaptation in S-GRPO~\citep{sgrpo}. Data-construction approaches include Uni-Edit~\citep{uniedit}, with 148K editing examples, and ThinkMorph~\citep{thinkmorph}, with approximately 24K reasoning traces. Specialized generation tasks include RecA~\citep{reca}, which primarily improves generation through image reconstruction, and UniMRG~\citep{unimrg}, which enhances perception using externally produced depth and segmentation targets. Self-judging methods use understanding-side feedback to guide generation, including GvU~\citep{gvu}, SRUM~\citep{SRUM}, and Internal Gap Self-Improvement~\citep{internalgap}. UniCorn~\citep{UniCorn} constructs supervision through self-play. UniRL~\citep{unirl} improves both tasks using self-generated images, but requires pre-constructed question-answer pairs as ground-truth supervision and applies a fixed round-trip via SFT and GRPO.

\section{Method}
\begin{figure}[t]
    \centering
    \includegraphics[width=\linewidth]{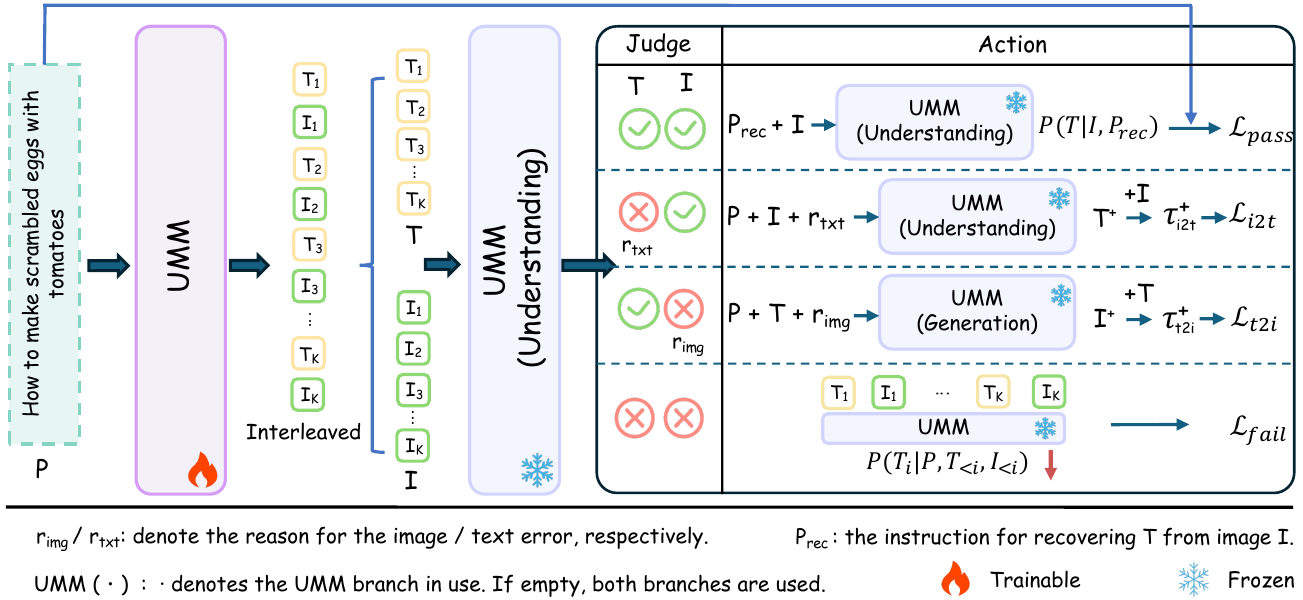}
    \caption{Overview of the Prism. Given a text prompt, Prism first uses the UMM to roll out an interleaved trajectory. It then uses the UMM in a judging role to assess the reliability of the text chain and image chain. Based on the resulting status, Prism selects the corresponding processing action and loss function. All UMM instances are initialized from the same base model. Only the rollout UMM is trainable, while the UMMs used for judging and action selection remain frozen.}
    \label{fig:architecture}
\end{figure}

Unlike prior post-training methods that rely on external supervision or a fixed supervision direction, Prism derives bidirectional supervision from self-generated interleaved trajectories. It constructs training targets according to the reliability of text and image chains, without ground-truth trajectories or external verifiers. As illustrated in Figure~\ref{fig:architecture}, given a task prompt, the model generates an interleaved trajectory. Frozen model copies then assess the text and image chains and map the trajectory to one of four training actions. Only the rollout model is updated using the resulting targets.

\subsection{Preliminary}
Given a text prompt $P$, the UMM generates an interleaved text-image trajectory:
\begin{equation}
\tau=(T_1,I_1,T_2,I_2,\ldots,T_K,I_K),
\label{eq:trajectory}
\end{equation}
where $T_i$ and $I_i$ denote the text and image segments at step $i$. For the interleaved tasks considered here, the trajectory has two modality-specific views, the complete text chain $T=(T_1,\ldots,T_K)$, representing understanding through planning and reasoning, and the complete image chain $I=(I_1,\ldots,I_K)$, representing generation through visual states and cross-step consistency. Both are projections of the same interleaved trajectory.

Standard UMM interleaved training models the text-image trajectory autoregressively:
\begin{equation}
p_{\theta}(\tau \mid P)
=
\prod_{i=1}^{K}
p_{\theta}(T_i \mid P, T_{<i}, I_{<i})
\,\,
p_{\theta}(I_i \mid P, T_{\le i}, I_{<i}).
\label{eq:interleaved_gen}
\end{equation}
The standard training loss for an interleaved trajectory is
\begin{equation}
\mathcal{L}_{\mathrm{interleave}}(\tau)
=
\sum_{i=1}^{K}
\left[
\lambda_{\mathrm{und}} \mathcal{L}_{\mathrm{und}}\!\left(f_{\theta}(\cdot \mid P,T_{<i},I_{<i}), T_i\right)
+
\lambda_{\mathrm{gen}} \mathcal{L}_{\mathrm{gen}}\!\left(f_{\theta}(\cdot \mid P,T_{\le i},I_{<i}), I_i\right)
\right],
\label{eq:interleave_loss}
\end{equation}
where $\mathcal{L}_{\mathrm{und}}$ is the token-level cross-entropy loss for text prediction, and $\mathcal{L}_{\mathrm{gen}}$ is the flow-matching loss for image generation~\citep{flowmatching}. $\lambda_{\mathrm{und}}$ and $\lambda_{\mathrm{gen}}$ are modality-balancing weights that may vary across architectures.

\subsection{Per-Modality Reliability Judgment}
\label{3.2}
The core premise of Prism is that, under the same task prompt $P$, the text chain $T$ and the image chain $I$ are two modality-specific expressions of the same underlying task content. Based on this premise, Prism judges the reliability of the complete text chain and complete image chain separately:
\begin{equation}
(v_{\mathrm{txt}},r_{\mathrm{txt}})=\mathcal{J}(P,T,q_{\mathrm{txt}}), \qquad
(v_{\mathrm{img}},r_{\mathrm{img}})=\mathcal{J}(P,I,q_{\mathrm{img}}),
\label{eq:judgment}
\end{equation}
where $\mathcal{J}$ is a frozen copy of the base model, and $q_{\mathrm{txt}}$ and $q_{\mathrm{img}}$ are review instructions. The indicators $v_{\mathrm{txt}},v_{\mathrm{img}}\in\{0,1\}$ denote the reliability of the understanding and generation sides, while $r_{\mathrm{txt}}$ and $r_{\mathrm{img}}$ specify detected errors. Prism judges the complete chains and uses the resulting states to select a training action, as illustrated in Figure~\ref{fig:architecture}.

\subsection{Reliability-Guided Supervision Construction}
\label{3.3}
Based on the judgments in Section~\ref{3.2}, Prism selects one of four actions for each trajectory: enhancing cross-modal consistency for reliable trajectories, rewriting an unreliable chain from the reliable side, or suppressing trajectories whose two chains are unreliable.

\noindent\textbf{Both pass, $v_{\mathrm{txt}} = 1, v_{\mathrm{img}} = 1$.}
When both chains are reliable, the trajectory provides a reliable interleaved training sample. Prism conditions text reconstruction on the image chain, reinforcing cross-modal consistency throughout the trajectory. Let $P_{\mathrm{rec}}$ denote the instruction for recovering $T$ from image chain $I$ under the original task. The objective is
\begin{equation}
\mathcal{L}_{\mathrm{pass}}
=
\mathcal{L}_{\mathrm{und}}
\bigl(f_{\theta}(T\mid P_{\mathrm{rec}},I),T\bigr).
\label{pass}
\end{equation}

\noindent\textbf{Text-to-image rewriting, $v_{\mathrm{txt}} = 1, v_{\mathrm{img}} = 0$.}
When the text chain is judged reliable but the image chain is not, Prism preserves $T$ and uses $P$, $T$, and $r_{\mathrm{img}}$ to generate a corrected image chain $I^+=(I^+_1,\ldots,I^+_K)$. The resulting corrected trajectory and training objective are
\begin{equation}
\tau^+_{\mathrm{t2i}}
=
(T_1,I^+_1,T_2,I^+_2,\ldots,T_K,I^+_K),
\qquad
\mathcal{L}_{\mathrm{t2i}}
=
\mathcal{L}_{\mathrm{interleave}}(\tau^+_{\mathrm{t2i}}).
\end{equation}

\noindent\textbf{Image-to-text rewriting, $v_{\mathrm{txt}} = 0, v_{\mathrm{img}} = 1$.}
When the image chain is judged reliable but the text chain is not, Prism preserves $I$ and uses $P$, $I$, and $r_{\mathrm{txt}}$ to generate a corrected text chain $T^+=(T^+_1,\ldots,T^+_K)$. The resulting corrected trajectory and training objective are
\begin{equation}
\tau^+_{\mathrm{i2t}}
=
(T^+_1,I_1,T^+_2,I_2,\ldots,T^+_K,I_K),
\qquad
\mathcal{L}_{\mathrm{i2t}}
=
\mathcal{L}_{\mathrm{interleave}}(\tau^+_{\mathrm{i2t}}).
\end{equation}

\noindent\textbf{Both fail, $v_{\mathrm{txt}} = 0, v_{\mathrm{img}} = 0$.}
When both chains are judged unreliable, Prism does not construct a positive target. Instead, it applies a token-level unlikelihood objective to the text tokens in the original trajectory, reducing the probability of reproducing its erroneous text patterns:
\begin{equation}
\mathcal{L}_{\mathrm{fail}}(\tau)
=
-\sum_{i=1}^{K}\sum_{n=1}^{|T_i|}
\log\left(1-p_{\theta}(t_{i,n}\mid P,T_{<i},I_{<i},t_{i,<n})\right),
\label{fail}
\end{equation}
where $T_i=(t_{i,1},\ldots,t_{i,|T_i|})$ denotes the $i$-th text segment and $t_{i,<n}$ denotes the tokens preceding $t_{i,n}$ within that segment. Unlike text, image generation is optimized through continuous flow matching, for which no direct counterpart of token-level unlikelihood is available. We therefore restrict negative supervision to the text tokens.

For the two rewriting cases, since the rewritten chain is newly generated and may fail to correct the detected errors, Prism verifies each rewritten trajectory before using it as a positive training target, retrying failed corrections and skipping trajectories that remain invalid. Appendix~\ref{rewritten} details the verification procedure. During training, only the rollout model is updated, while judgment, rewriting, and verification use frozen copies of the base checkpoint to prevent co-evolution with the rollout model. Jointly updating them leads to training collapse, as shown in Appendix~\ref{app:freezing_strategy}.

\subsection{Training Objective}
Each trajectory contributes to one of the four objectives according to its reliability state. So the overall objective is
\begin{equation}
\mathcal{L}_{\mathrm{total}}
=
\lambda_{\mathrm{pass}}\mathcal{L}_{\mathrm{pass}}
+
\mathcal{L}_{\mathrm{t2i}}
+
\mathcal{L}_{\mathrm{i2t}}
+
\lambda_{\mathrm{fail}}\mathcal{L}_{\mathrm{fail}},
\label{eq:total}
\end{equation}
where $\lambda_{\mathrm{pass}}$, $\lambda_{\mathrm{i2t}}$, and $\lambda_{\mathrm{fail}}$ weight the both-pass, image-to-text rewriting, and both-fail objectives, respectively. The text-to-image rewriting objective uses unit weight.

\section{Experiments}

\subsection{Experimental Setup}
\label{setup}
\textbf{Data.} We use GPT-5.5 to generate 2.4K task prompt seeds spanning tutorials, stories, spatial reasoning, and other task types. These prompts provide task descriptions only. Prism generates interleaved trajectories online and constructs training targets from the model's own rollouts.

\textbf{Evaluation.} We evaluate image generation on GenEval~\citep{geneval} and DPG-Bench~\citep{dpg}, multimodal understanding on MME~\citep{mme}, MMBench~\citep{mmb}, MMMU~\citep{mmmu}, MMStar~\citep{mmstar}, MMVP~\citep{mmvp}, and POPE~\citep{pope}, and interleaved generation on OpenING~\citep{opening}. We follow the official evaluation settings for all benchmarks and report the multiple-choice accuracy for MMMU.

\textbf{Training.} Since some UMMs lack interleaved text-image generation support or publicly available model weights, we select SenseNova-U1~\citep{sensenova} and BAGEL~\citep{bagel}, which satisfy both requirements. The main experiments use a learning rate of 3e-5, with $\lambda_{\mathrm{pass}}=0.5$, and $\lambda_{\mathrm{fail}}=0.005$. The text-to-image rewriting objective uses unit weight. Training is performed on 16 A800 GPUs for approximately 24 hours.

For more details, please refer to the \textbf{Appendix~\ref{training and evaluation}}.

\subsection{Prism Simultaneously Improves Understanding and Generation}
We evaluate Prism on SenseNova-U1 and BAGEL across image generation and multimodal understanding benchmarks. Tables~\ref{tab:generation} and \ref{tab:understanding} show consistent gains on both models.

On SenseNova-U1, Prism improves GenEval by 3.4 points and DPG by 1.2 points, with notable gains in counting, two-object generation, and attribute binding. On understanding benchmarks, it improves MME by 29.0 points and MMVP by 4.0 points, with gains on the other benchmarks as well. BAGEL shows the same overall trend on the benchmarks reported in Tables 1 and 2, providing evidence that the method can transfer across the two evaluated UMM backbones.

\begin{table}[t]
\caption{Prism improves image generation performance.}
\label{tab:generation}
\centering
\setlength{\tabcolsep}{4pt}
\begin{tabular}{c c ccccccc c}
\multirow{2}{*}{\textbf{Model}} & \multirow{2}{*}{\textbf{Prism}} & \multicolumn{7}{c}{\textbf{GenEval}} & \multirow{2}{*}{\textbf{DPG}} \\
\cmidrule(lr){3-9}
& & \textbf{Single} & \textbf{Two} & \textbf{Count} & \textbf{Color} & \textbf{Position} & \textbf{Attri.} & \textbf{Overall} & \\
\midrule

\multirow{2}{*}{\cellm{SenseNo\\va-U1}}
& \cellr{\xmark}
& \cellg{100.0}{}
& \cellg{92.9}{}
& \cellg{75.6}{}
& \cellg{91.7}{}
& \cellg{85.5}{}
& \cellg{68.7}{}
& \cellg{85.8}{}
& \celld{86.1}{} \\
{}
& \cellcolor{green!5}\cellr{\cmark}
& \cellcolor{green!5}\cellg{\textbf{100.0}}{\gain{0.0}}
& \cellcolor{green!5}\cellg{\textbf{96.7}}{\gain{+3.8}}
& \cellcolor{green!5}\cellg{\textbf{87.2}}{\gain{+11.6}}
& \cellcolor{green!5}\cellg{\textbf{89.6}}{\gain{-2.1}}
& \cellcolor{green!5}\cellg{\textbf{86.8}}{\gain{+1.3}}
& \cellcolor{green!5}\cellg{\textbf{75.0}}{\gain{+6.3}}
& \cellcolor{green!5}\cellg{\textbf{89.2}}{\gain{+3.4}}
& \cellcolor{green!5}\celld{\textbf{87.3}}{\gain{+1.2}} \\
\midrule

\multirow{2}{*}{\cellm{BAGEL}}
& \cellr{\xmark}
& \cellg{99.1}{}
& \cellg{93.0}{}
& \cellg{76.5}{}
& \cellg{86.0}{}
& \cellg{47.2}{}
& \cellg{57.0}{}
& \cellg{76.5}{}
& \celld{84.0}{} \\
{}
& \cellcolor{green!5}\cellr{\cmark}
& \cellcolor{green!5}\cellg{\textbf{99.4}}{\gain{+0.3}}
& \cellcolor{green!5}\cellg{\textbf{96.5}}{\gain{+3.5}}
& \cellcolor{green!5}\cellg{\textbf{76.9}}{\gain{+0.4}}
& \cellcolor{green!5}\cellg{\textbf{86.5}}{\gain{+0.5}}
& \cellcolor{green!5}\cellg{\textbf{48.0}}{\gain{+0.8}}
& \cellcolor{green!5}\cellg{\textbf{62.5}}{\gain{+5.5}}
& \cellcolor{green!5}\cellg{\textbf{78.3}}{\gain{+1.8}}
& \cellcolor{green!5}\celld{\textbf{85.0}}{\gain{+1.0}} \\
\end{tabular}
\end{table}

\begin{table}[t]
\caption{Prism improves multimodal understanding performance. We evaluate the effectiveness of Prism across understanding benchmarks covering general knowledge, subject knowledge, and others.}
\label{tab:understanding}
\centering
\setlength{\tabcolsep}{6pt}
\begin{tabular}{c c @{\hspace{9pt}} c @{\hspace{9pt}} c c c c c}
\textbf{Model} & \textbf{Prism} & \textbf{MME-P} & \textbf{MME-C} & \textbf{MMBench} & \textbf{MMMU} & \textbf{MMStar} & \textbf{MMVP} \\
\midrule

\multirow{2}{*}{SenseNova-U1}
& \xmark
& \cellp{1699.7}{}
& \cellc{582.9}{}
& \cells{86.1}{}
& \cells{59.0}{}
& \cells{65.1}{}
& \cells{58.0}{} \\
{}
& \cellcolor{green!5}\cmark
& \cellcolor{green!5}\cellp{\textbf{1713.4}}{\gain{+13.7}}
& \cellcolor{green!5}\cellc{\textbf{598.2}}{\gain{+15.3}}   
& \cellcolor{green!5}\cells{\textbf{87.3}}{\gain{+1.2}}
& \cellcolor{green!5}\cells{\textbf{60.6}}{\gain{+1.6}}    
& \cellcolor{green!5}\cells{\textbf{66.5}}{\gain{+1.4}}
& \cellcolor{green!5}\cells{\textbf{62.0}}{\gain{+4.0}} \\
\midrule

\multirow{2}{*}{BAGEL}
& \xmark
& \cellp{1681.0}{}
& \cellc{679.6}{}
& \cells{84.6}{}
& \cells{53.7}{}
& \cells{65.0}{}
& \cells{\textbf{69.3}}{} \\
{}
& \cellcolor{green!5}\cmark
& \cellcolor{green!5}\cellp{\textbf{1685.2}}{\gain{+4.2}} 
& \cellcolor{green!5}\cellc{\textbf{685.4}}{\gain{+5.8}}   
& \cellcolor{green!5}\cells{\textbf{88.4}}{\gain{+3.8}}
& \cellcolor{green!5}\cells{\textbf{55.0}}{\gain{+1.3}}
& \cellcolor{green!5}\cells{\textbf{69.6}}{\gain{+4.6}}
& \cellcolor{green!5}\cells{70.0}{\gain{+0.7}} \\          
\end{tabular}
\end{table}

\begin{table}[!t]
\centering
\caption{\textbf{OpenING pairwise evaluation.}
\textbf{Left:} Win rates under different tie-handling settings.
\textbf{Right:} Zero-image rates, i.e., outputs with no generated image. Lower is better.}
\label{tab:tie_comparison}
\setlength{\tabcolsep}{4pt}
\small
\begin{minipage}[t]{0.57\linewidth}
\centering
\vspace{0pt}
\begin{tabular}{lcccc}
\toprule
\textbf{Method} & \textbf{FDT} & \textbf{w/o Tie} & \textbf{w/ Tie (0)} & \textbf{w/ Tie (.5)} \\
\midrule
\rowcolor{green!5}
SenseNova-U1 + Prism & \textbf{74.93\%} & \textbf{76.65\%} & \textbf{66.79\%} & \textbf{73.22\%} \\
\rowcolor{prismblue}
SenseNova-U1 & 72.03\% & 72.91\% & 62.55\% & 69.66\% \\
GPT-4o+DALL-E3 & 63.79\% & 66.92\% & 56.36\% & 64.25\% \\
Human & 61.10\% & 64.66\% & 51.90\% & 61.76\% \\
Gemini1.5+Flux & 38.35\% & 39.04\% & 31.23\% & 41.24\% \\
\rowcolor{green!5}
BAGEL + Prism & 34.70\% & 31.66\% & 25.80\% & 35.05\% \\
\rowcolor{prismblue}
BAGEL & 30.84\% & 27.94\% & 23.47\% & 31.47\% \\
SEED-X & 21.27\% & 17.42\% & 15.92\% & 20.23\% \\
\bottomrule
\end{tabular}
\end{minipage}
\hfill
\begin{minipage}[t]{0.36\linewidth}
\centering
\vspace{-2pt}
\includegraphics[width=\linewidth]{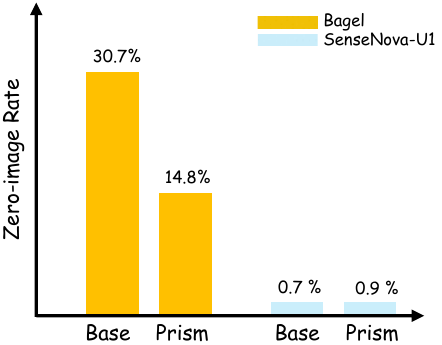}
\end{minipage}
\end{table}

\begin{figure}[!t]
    \centering
    \includegraphics[width=\linewidth]{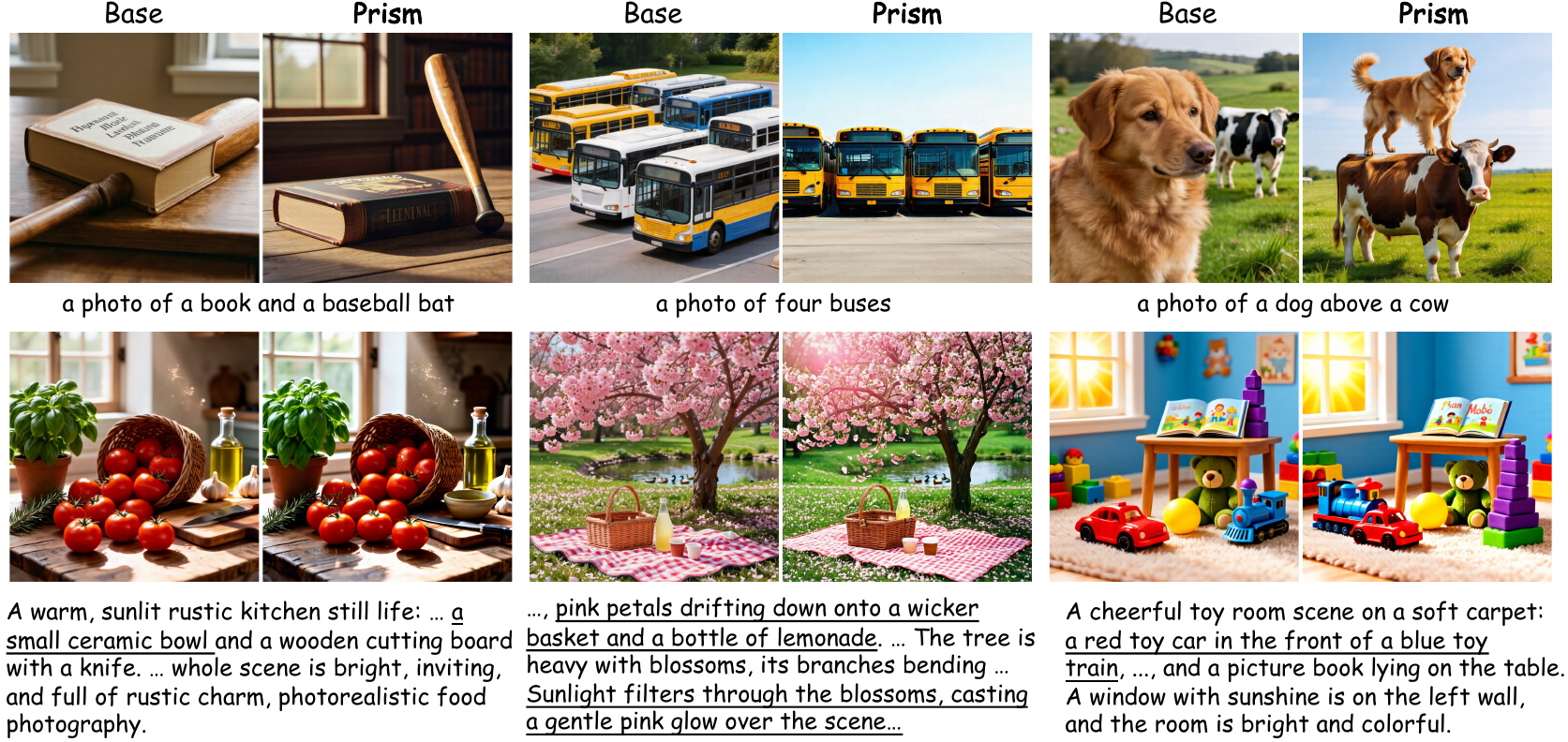}
    \caption{Improved generation. We use SenseNova-U1 as the baseline model. The model trained with Prism shows more accurate adherence to prompt, including spatial relations, object counts, and style. The underlined text highlights the generation differences under complex instructions.}
    \label{fig:generation_comparsion}
\end{figure}

\begin{table}[!t]
\caption{Compared with similar methods on BAGEL, our method can simultaneously improve both understanding and generation, demonstrating bidirectional enhancement capability.}
\label{tab:comparison}
\centering
\setlength{\tabcolsep}{6pt}
\begin{tabular}{c @{\hspace{9pt}} c @{\hspace{9pt}} c c c c c c}
\textbf{Model} & \textbf{MME} & \textbf{MMBench} & \textbf{MMMU} & \textbf{MMStar} & \textbf{MMVP} & \textbf{GenEval} & \textbf{DPG} \\
\midrule

BAGEL
& \cellc{2361}{}
& \cells{84.6}{}
& \cells{53.7}{}
& \cells{65.0}{}
& \cellg{69.3}{}
& \cellg{76.5}{}
& \celld{84.0}{} \\
\midrule

+ SRUM
& \cellc{2350}{\gain{-11}}
& \cells{84.4}{\gain{-0.2}}
& \cells{53.4}{\gain{-0.3}}
& \cells{64.7}{\gain{-0.3}}
& \cellg{70.0}{\gain{+0.7}}
& \cellg{77.4}{\gain{+0.9}}
& \celld{85.2}{\gain{+1.2}} \\
\midrule

+ UniCorn
& \cellc{2337}{\gain{-24}}
& \cells{84.1}{\gain{-0.5}}
& \cells{54.5}{\gain{+0.8}}
& \cells{65.0}{\gain{+0.0}}
& \cellg{70.0}{\gain{+0.7}}
& \cellg{80.3}{\gain{+3.8}}
& \celld{86.2}{\gain{+2.2}} \\
\midrule

\rowcolor{green!5} + Prism
& \cellc{2371}{\gain{+10}}
& \cells{88.4}{\gain{+3.8}}
& \cells{55.0}{\gain{+1.3}}
& \cells{69.6}{\gain{+4.6}}
& \cellg{70.0}{\gain{+0.7}}
& \cellg{78.3}{\gain{+1.8}}
& \celld{85.0}{\gain{+1.0}} \\
\end{tabular}
\end{table}

\subsection{Prism Improves Interleaved Generation}
We evaluate Prism on OpenING~\cite{opening}, which measures open-ended interleaved image--text generation with pairwise win rates. Table~\ref{tab:tie_comparison} reports the win rates and zero-image rates.

Prism improves SenseNova-U1 by 2.90 to 4.24 points and BAGEL by 2.33 to 3.86 points across all tie-handling settings. Among the compared entries, SenseNova-U1 with Prism achieves the highest win rate in all four settings. Prism also reduces the zero-image rate of BAGEL from 30.7\% to 14.8\%, indicating improved ability to generate interleaved trajectories. SenseNova-U1 already has a near-zero zero-image rate of 0.7\% before training, and Prism preserves this capability with a rate of 0.9\%. Overall, these results show that Prism improves interleaved generation across two UMMs.

\subsection{Comparison with Related Post-Training Methods}
We compare Prism with SRUM and UniCorn on BAGEL in Table~\ref{tab:comparison}. SRUM and UniCorn both improve GenEval and DPG, but reduce MME and MMBench. On other understanding benchmarks, some scores stay flat or drop. This shows that they do not achieve consistent gains in both generation and multimodal understanding. By contrast, Prism improves MME by 10 points, MMBench by 3.8 points, MMMU by 1.3 points, and MMStar by 4.6 points, while maintaining comparable MMVP performance. It also improves GenEval by 1.8 points and DPG by 1.0 point. Results show the four losses contribute distinctly to understanding and generation.

\subsection{Ablation Experiment}
To ablate each loss, we train trajectories assigned to that loss on their unmodified rollouts with $\mathcal{L}_{\mathrm{interleave}}$, leaving the other losses unchanged. Table~\ref{tab:ablation} shows that replacing $\mathcal{L}_{\mathrm{t2i}}$ reduces GenEval by 1.2 points, while replacing $\mathcal{L}_{\mathrm{i2t}}$ reduces MMVP by 5.3 points. Replacing $\mathcal{L}_{\mathrm{pass}}$ slightly improves GenEval but lowers four of five understanding metrics. Replacing $\mathcal{L}_{\mathrm{fail}}$ lowers both generation metrics and four of five understanding metrics. The four losses play distinct roles in improving understanding, generation, and cross-modal consistency.

\begin{figure}[!t]
    \centering
    \includegraphics[width=\linewidth]{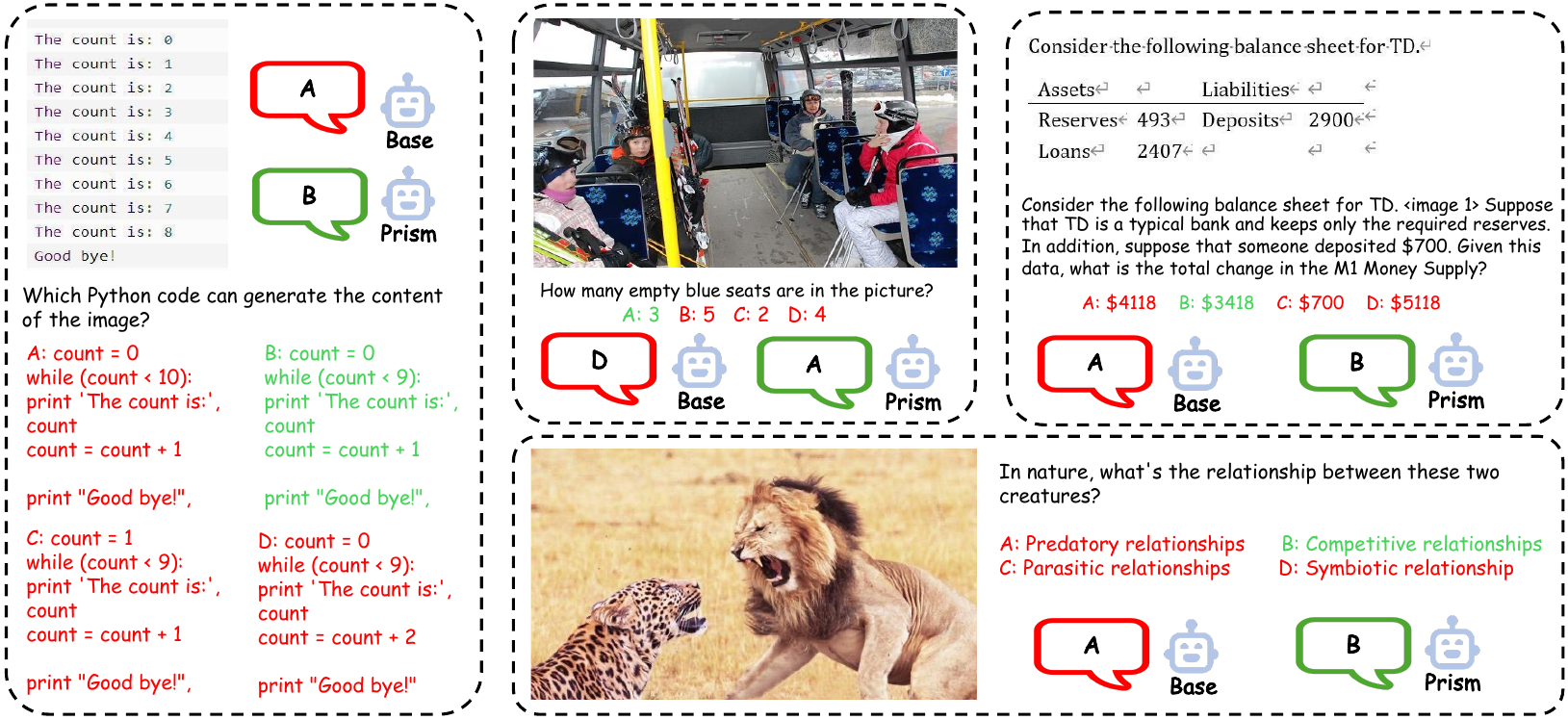}
\caption{Improved understanding. We use SenseNova-U1 as the baseline model. The model trained with Prism demonstrates more accurate responses on diverse multimodal multiple-choice tasks, including code reasoning, visual counting, numerical reasoning, and commonsense understanding. For more examples, please refer to the Appendix~\ref{understanding examples}}
    \label{fig:understanding_comparsion}
\end{figure}

\begin{table}[!t]
\caption{Loss ablations on SenseNova-U1. Dashes denote the base model without post-training. Crosses denote replacement with $\mathcal{L}_{\mathrm{interleave}}$ on the corresponding unmodified rollouts.}
\label{tab:ablation}
\centering
\setlength{\tabcolsep}{2pt}
\footnotesize
\begin{tabular}{
  c @{\hspace{2pt}} c @{\hspace{2pt}} c @{\hspace{2pt}} c
  @{\hspace{8pt}}
  c
  @{}
  c c c c c c
}
\multicolumn{4}{c}{\textbf{Loss Terms}} & \multicolumn{5}{c}{\textbf{Understanding}} & \multicolumn{2}{c}{\textbf{Generation}} \\
\cmidrule(lr){1-4} \cmidrule(lr){5-9} \cmidrule(lr){10-11}
$\mathcal{L}_{\mathrm{pass}}$ & $\mathcal{L}_{\mathrm{t2i}}$ & $\mathcal{L}_{\mathrm{i2t}}$ & $\mathcal{L}_{\mathrm{fail}}$ & \textbf{MME} & \textbf{MMBench} & \textbf{MMMU} & \textbf{MMStar} & \textbf{MMVP} & \textbf{GenEval} & \textbf{DPG} \\
\midrule

- & - & - & -
& \ablmme{2283}{}
& \abldec{86.1}{}
& \abldec{59.0}{}
& \abldec{65.1}{}
& \abldec{58.0}{}
& \ablge{85.8}{}
& \abldpg{86.1}{} \\
\midrule

\xmark & \cmark & \cmark & \cmark
& \ablmme{2299}{\gain{+16}}
& \abldec{86.6}{\gain{+0.5}}
& \abldec{61.0}{\gain{+2.0}}
& \abldec{65.3}{\gain{+0.2}}
& \abldec{60.0}{\gain{+2.0}}
& \ablge{\textbf{89.4}}{\gain{+3.6}}
& \abldpg{87.2}{\gain{+1.1}} \\

\cmark & \xmark & \cmark & \cmark
& \ablmme{2294}{\gain{+11}}
& \abldec{\textbf{87.4}}{\gain{+1.3}}
& \abldec{60.1}{\gain{+1.1}}
& \abldec{64.9}{\gain{-0.2}}
& \abldec{59.3}{\gain{+1.3}}
& \ablge{88.0}{\gain{+2.2}}
& \abldpg{87.0}{\gain{+0.9}} \\

\cmark & \cmark & \xmark & \cmark
& \ablmme{2299}{\gain{+16}}
& \abldec{86.6}{\gain{+0.5}}
& \abldec{60.1}{\gain{+1.1}}
& \abldec{65.6}{\gain{+0.5}}
& \abldec{56.7}{\gain{-1.3}}
& \ablge{88.8}{\gain{+3.0}}
& \abldpg{\textbf{87.3}}{\gain{+1.2}} \\

\cmark & \cmark & \cmark & \xmark
& \ablmme{2310}{\gain{+27}}
& \abldec{86.8}{\gain{+0.7}}
& \abldec{\textbf{61.3}}{\gain{+2.3}}
& \abldec{65.3}{\gain{+0.2}}
& \abldec{60.0}{\gain{+2.0}}
& \ablge{88.7}{\gain{+2.9}}
& \abldpg{87.1}{\gain{+1.0}} \\

\cmark & \cmark & \cmark & \cmark
& \ablmme{\textbf{2312}}{\gain{+29}}
& \abldec{87.3}{\gain{+1.2}}
& \abldec{60.6}{\gain{+1.6}}
& \abldec{\textbf{66.5}}{\gain{+1.4}}
& \abldec{\textbf{62.0}}{\gain{+4.0}}
& \ablge{89.2}{\gain{+3.4}}
& \abldpg{\textbf{87.3}}{\gain{+1.2}} \\
\end{tabular}
\end{table}

\begin{table*}[!t]
\centering
\caption{Analysis of reliability-guided target construction on SenseNova-U1.}
\label{tab:overall_benchmarks}
\setlength{\tabcolsep}{5pt}
\begin{tabular}{lccccccc}
\textbf{Setting} & \textbf{GenEval} & \textbf{DPG} & \textbf{MME} &
\textbf{MMB} & \textbf{MMMU} & \textbf{MMStar} & \textbf{MMVP} \\
\midrule
Base Model
& 85.8 & 86.1 & 2282.6 & 86.1 & 59.0 & 65.1 & 58.0 \\
Base w/ Self-Training
& 87.7 & 86.6 & 2276.3 & 85.5 & 60.0 & 64.7 & 58.0 \\
Prism w/o Rewriting
& 88.4 & 86.5 & 2300.6 & 86.8 & \textbf{61.0} & 65.6 & 60.7 \\
Prism w/ Random Routing
& 87.6 & 86.5 & 2297.5 & 86.5 & 59.9 & 65.5 & 59.3 \\
\rowcolor{green!5}
Prism (Ours)
& \textbf{89.2} & \textbf{87.3} & \textbf{2311.6} &
\textbf{87.3} & 60.6 & \textbf{66.5} & \textbf{62.0} \\
\end{tabular}
\end{table*}

\subsection{Analysis of Reliability-Guided Target Construction}
Table~\ref{tab:overall_benchmarks} compares Prism with three controls. Base w/ Self-Training applies $\mathcal{L}_{\mathrm{interleave}}$ to all unmodified rollouts. Prism w/o Rewriting keeps judge-based routing but applies $\mathcal{L}_{\mathrm{interleave}}$ to the original rollouts assigned to t2i and i2t, leaving pass and fail unchanged. Prism w/ Random Routing shuffles action assignments within each batch while preserving their counts. Full Prism outperforms self-training and random routing on all seven metrics, and w/o rewriting on six. These results support the contributions of reliability-guided routing and rewritten targets. More appears in Appendix~\ref{app:detailed reliability}.

\subsection{More Results}
\begin{figure}[t]
    \centering
    \includegraphics[width=\linewidth]{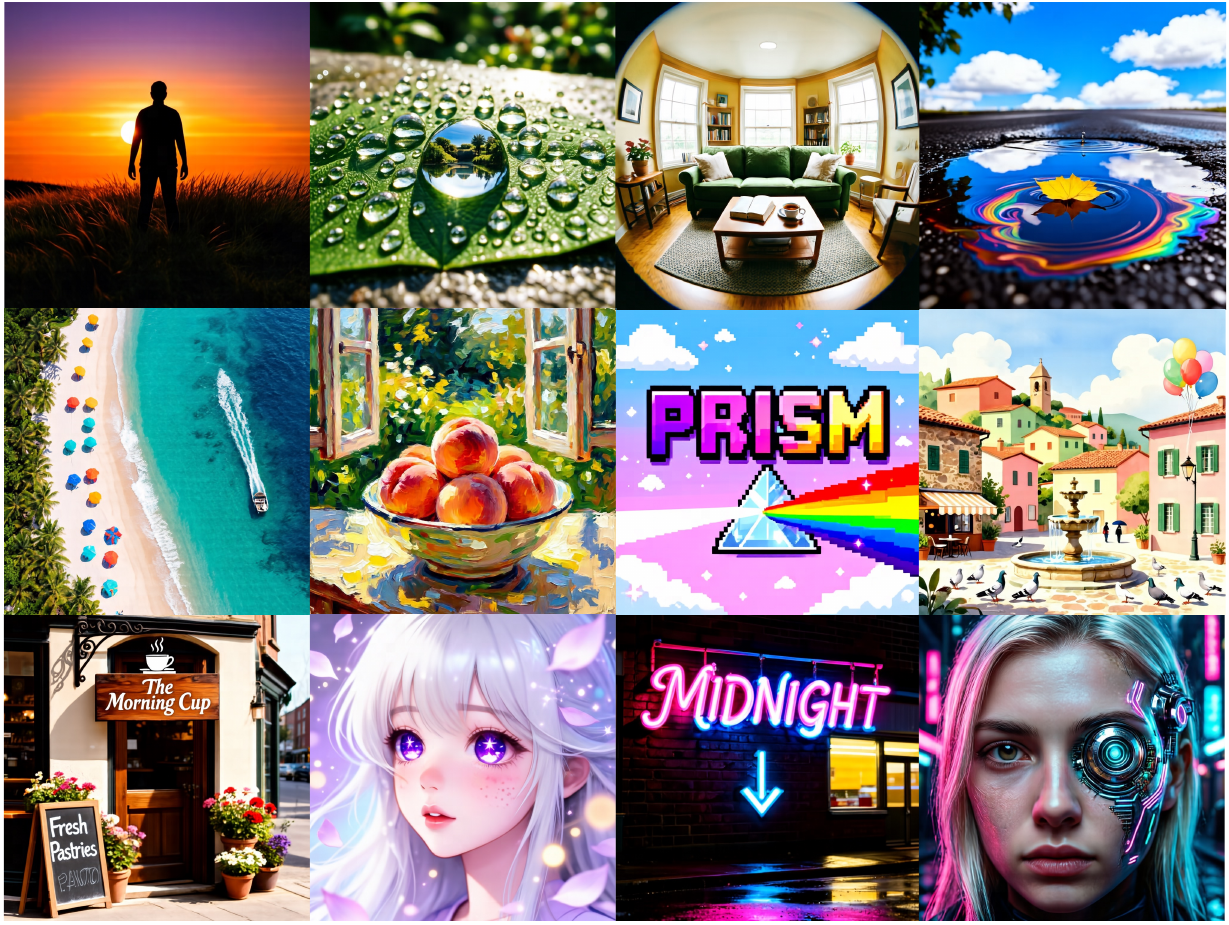}
    \caption{Qualitative examples generated by the model trained with Prism. The figure illustrates a diverse set of outputs, including realistic photography and artistic rendering, among others.}
    \label{fig:showimages}
\end{figure}

\textbf{Qualitative.} Figure~\ref{fig:generation_comparsion} compares SenseNova-U1 with its Prism-trained variant. Prism more faithfully follows prompts involving object composition, counting, spatial relations, and detailed scene descriptions. Figure~\ref{fig:understanding_comparsion} further shows improved responses on code reasoning, visual counting, document-based numerical reasoning, and relation understanding. These examples complement the quantitative results on generation and multimodal understanding. More examples with diverse styles and quantities are provided in Figure~\ref{fig:showimages} and Appendix~\ref{genereation examples}.

\noindent
\begin{minipage}[t]{0.58\linewidth}
\vspace{0pt}
\textbf{Hallucination analysis.}
We further evaluate Prism on POPE. As shown in Table~\ref{tab:pope}, Prism improves Accuracy and F1 from 85.2 to 88.0 and from 84.9 to 86.7 on BAGEL, respectively. On SenseNova-U1, it improves Accuracy from 85.6 to 87.5 and F1 from 84.0 to 86.6. These results suggest improved visual grounding and fewer hallucination-related errors.
\end{minipage}\hfill
\begin{minipage}[t]{0.38\linewidth}
\vspace{-3pt}
\centering
\captionsetup{aboveskip=0pt,belowskip=3pt}
\captionof{table}{POPE benchmark.}
\label{tab:pope}
\setlength{\tabcolsep}{4pt}
\footnotesize
\begin{tabular}{c c c c}
\toprule
\textbf{Model} & \textbf{Prism} & \textbf{Acc.} & \textbf{F1} \\
\midrule
\multirow{2}{*}{BAGEL}
& \xmark & \cells{85.2}{} & \cells{84.9}{} \\
{}
& \cellcolor{green!5}\cmark
& \cellcolor{green!2}\cells{88.0}{\gain{+2.8}}
& \cellcolor{green!5}\cells{86.7}{\gain{+1.8}} \\
\midrule
\multirow{2}{*}{SenseNova-U1}
& \xmark & \cells{85.6}{} & \cells{84.0}{} \\
{}
& \cellcolor{green!5}\cmark
& \cellcolor{green!2}\cells{87.5}{\gain{+1.9}}
& \cellcolor{green!5}\cells{86.6}{\gain{+2.6}} \\
\bottomrule
\end{tabular}
\end{minipage}

\section{Conclusion and Limitations}
We propose Prism, a bidirectional self-teaching method that derives targets through text and image reliability. It improves generation, understanding and interleaved generation on UMMs.

Despite these gains, Prism has three limitations. It targets interleaved text-image tasks with externally provided prompt seeds, and applies negative supervision to the complete text chain when both chains fail. Future work will extend Prism to broader multimodal tasks, fully self-sampled training, and localized negative supervision.

\bibliography{iclr2027_conference}
\bibliographystyle{iclr2027_conference}

\newpage
\appendix
\section{Training and Evaluation Details}
\label{training and evaluation}
\subsection{Implementation Details and Hyperparameters}

We train all models at a resolution of $1024\times1024$ for one epoch. For the interleaved objective, the cross-entropy and flow-matching loss weights are set to $0.1:1$ for SenseNova-U1 and $0.25:1$ for BAGEL. We use AdamW with a learning rate of $3\times10^{-5}$ and an effective batch size of 64. The reported training time includes online rollout generation, reliability judgment, rewriting, and verification.

For rollout and image generation, we use 50 diffusion steps with global normalization. The text and image classifier-free guidance scales are set to 4.0 and 1.0, respectively. Each prompt is assigned a fixed random seed shared across compared models. 

\subsection{OpenING Arena Protocol}

We follow the official pairwise arena protocol of OpenING~\citep{opening}. The evaluation set contains 2,160 samples. For each sample, two distinct entries are paired through circular rotation and random permutation, and their presentation order is randomized to mitigate position bias.

The compact comparison in the main paper contains eight entries and 4,320 pairwise battles, with each entry participating in 1,080 battles on average. The full leaderboard in Table~\ref{tab:opening_leaderboard} contains 15 entries, including our Prism-trained models, their corresponding base models, and 11 official OpenING reference outputs. It contains 4,143 valid pairwise battles. The number of battles per entry varies between 507 and 615, as reported in the final column of Table~\ref{tab:opening_leaderboard}. Since the compact comparison and the full leaderboard are evaluated in separate arenas, their win rates are arena-relative and are not directly comparable.

We use the official \textbf{IntJudge-8B} checkpoint and decoding configuration. For each battle, IntJudge outputs one of four decisions: \textit{A}, \textit{B}, \textit{Tie(A)}, or \textit{Tie(B)}. We report four tie-handling metrics:
\begin{itemize}
    \item \textbf{FDT:} \textit{Tie(A)} is counted as a win for A, and \textit{Tie(B)} is counted as a win for B.
    \item \textbf{w/o Tie:} only battles with an explicit winner are counted.
    \item \textbf{w/ Tie (0):} ties contribute zero wins to both models.
    \item \textbf{w/ Tie (.5):} ties contribute 0.5 wins to both models.
\end{itemize}

We use \textit{w/ Tie (.5)} as the main reporting protocol. The \textit{Human} entry uses the human-authored interleaved image--text sequences released with OpenING and participates in the same arena protocol as model outputs. Under the \textit{w/o Tie} protocol, SenseNova-U1 with Prism achieves an 86.61\% win rate over 478 decisive outcomes from 537 battles, exceeding the base model by 3.55 points and GPT-4o+DALL-E3 by 5.23 points.

\begin{table*}[!h]
\caption{Full OpenING arena leaderboard. We report pairwise win rates under four tie-handling protocols. The final column reports decisive battles over total battles for each entry.}
\label{tab:opening_leaderboard}
\centering
\small
\setlength{\tabcolsep}{8pt}
\begin{tabular}{clccccc}
\textbf{Rank} & \textbf{Model} & \textbf{FDT} & \textbf{w/o Tie} & \textbf{w/ Tie (0)} & \textbf{w/ Tie (.5)} & \textbf{Decisive / Total} \\
\midrule
1  & SenseNova-U1 + Prism & \textbf{84.73\%} & \textbf{86.61\%} & \textbf{77.09\%} & \textbf{82.59\%} & 478 / 537 \\
2  & SenseNova-U1         & 82.33\% & 83.06\% & 71.91\% & 78.62\% & 490 / 566 \\
3  & GPT-4o+DALL-E3       & 77.78\% & 81.38\% & 69.79\% & 76.91\% & 494 / 576 \\
4  & Human                & 76.88\% & 82.92\% & 67.50\% & 76.80\% & 486 / 597 \\
5  & Gemini1.5+Flux       & 62.19\% & 64.16\% & 50.36\% & 61.11\% & 438 / 558 \\
6  & BAGEL + Prism        & 54.38\% & 54.79\% & 42.93\% & 53.76\% & 438 / 559 \\
7  & SEED-LLaMA           & 43.32\% & 47.04\% & 35.92\% & 47.74\% & 423 / 554 \\
8  & Anole                & 48.66\% & 44.18\% & 35.50\% & 45.32\% & 421 / 524 \\
9  & BAGEL-7B-MoT         & 43.76\% & 40.89\% & 33.59\% & 42.51\% & 428 / 521 \\
10 & Emu3                 & 42.59\% & 37.56\% & 30.99\% & 39.73\% & 434 / 526 \\
11 & Emu2                 & 36.32\% & 36.20\% & 27.54\% & 39.50\% & 442 / 581 \\
12 & MiniGPT-5            & 26.02\% & 27.08\% & 21.14\% & 32.11\% & 480 / 615 \\
13 & SEED-X               & 30.84\% & 29.17\% & 27.48\% & 30.37\% & 504 / 535 \\
14 & MiniGPT-5mmd         & 20.57\% & 17.36\% & 14.91\% & 21.98\% & 455 / 530 \\
15 & NExT-GPT             & 14.20\% & 10.33\% & 9.27\% & 14.40\% & 455 / 507 \\
\end{tabular}
\end{table*}

\subsection{Image Generation Evaluation}

We evaluate image generation with the official data and scoring pipelines of GenEval~\citep{geneval} and DPG-Bench~\citep{dpg}. GenEval contains 553 prompts, with four images generated per prompt. It evaluates six categories using Mask2Former and CLIP, and reports their equal-weighted average. DPG-Bench contains 1,065 dense prompts and uses mPLUG-VQA for evaluation.

All images are generated at $1024\times1024$ using the configuration above. For DPG-Bench, we generate four images per prompt and arrange them into a $2\times2$ grid. The official evaluator crops and scores each image at the native $1024$ resolution. All comparisons with the base models use identical generation settings.

\subsection{Multimodal Understanding Evaluation}

We evaluate multimodal understanding using the official data, prompts, and evaluation scripts of MME, MMBench, MMMU, MMStar, POPE, and MMVP. No task-specific prompts are added.

MME contains 14 tasks, including 10 perception tasks with a maximum score of 2,000 and four cognition tasks with a maximum score of 800. We report the total score in the main paper. MMBench is evaluated on its 1,176-example development split with the official option-rotation protocol. MMMU is evaluated on its 900-example validation split, including 847 multiple-choice and 53 open-ended questions. We report multiple-choice accuracy. MMStar contains 1,500 examples. POPE includes 9,000 questions from the random, popular, and adversarial settings, for which we report Accuracy and F1. MMVP contains 300 questions organized into 150 paired examples, and is scored only when both questions in a pair are answered correctly.

All understanding benchmarks are evaluated using a unified decoding configuration and the official aggregation scripts.

\section{Additional Capability Analysis}

Prism constructs supervision from the consistency between the complete text and image chains of self-generated interleaved trajectories. Its training objectives therefore emphasize planning, visual state progression, and cross-modal consistency, rather than instruction following for single-image generation or image editing. This appendix reports results on benchmarks that fall outside this scope. All benchmarks are evaluated with the official protocols and judge models released by the benchmark authors.

\begin{table*}[t]
\caption{OneIG-Bench results. Each model generates four images per prompt on 1{,}120 English prompts. The five dimensions are scored with dedicated scorers and the benchmark provides no aggregate score. The better value is in bold.}
\label{tab:oneig}
\centering
\setlength{\tabcolsep}{8pt}
\begin{tabular}{c c ccccc}
\multirow{2}{*}{\textbf{Model}} & \multirow{2}{*}{\textbf{Prism}} & \multicolumn{5}{c}{\textbf{OneIG-Bench}} \\
\cmidrule(lr){3-7}
& & \textbf{Alignment} & \textbf{Text} & \textbf{Style} & \textbf{Reasoning} & \textbf{Diversity} \\
\midrule

\multirow{2}{*}{SenseNova-U1}
& \cellr{\xmark}
& \cellg{\textbf{0.873}}{}
& \cellg{\textbf{0.942}}{}
& \cellg{0.405}{}
& \cellg{0.326}{}
& \cellg{0.170}{} \\
{}
& \cellr{\cmark}
& \cellg{0.869}{}
& \cellg{0.937}{}
& \cellg{\textbf{0.410}}{}
& \cellg{\textbf{0.330}}{}
& \cellg{\textbf{0.183}}{} \\
\midrule

\multirow{2}{*}{\cellm{BAGEL}}
& \cellr{\xmark}
& \cellg{\textbf{0.813}}{}
& \cellg{\textbf{0.366}}{}
& \cellg{0.395}{}
& \cellg{0.189}{}
& \cellg{0.176}{} \\
{}
& \cellr{\cmark}
& \cellg{0.808}{}
& \cellg{0.339}{}
& \cellg{0.395}{}
& \cellg{0.189}{}
& \cellg{\textbf{0.179}}{} \\
\end{tabular}
\end{table*}

\noindent\textbf{OneIG-Bench.} OneIG-Bench evaluates text-to-image generation on 1{,}120 English prompts with four images generated per prompt \citep{oneig}. As shown in Table~\ref{tab:oneig}, the results are mixed but close to the base models. On SenseNova-U1, Prism improves Diversity by 0.013, Style by 0.005, and Reasoning by 0.004, while Alignment and Text decrease slightly by about 0.005. The improved dimensions are consistent with training on trajectories that require coordinated textual plans and visual states. On BAGEL, Style, Reasoning, and Diversity are unchanged, Alignment decreases slightly, and Text shows the largest decrease of 0.027. Text rendering is the weakest dimension of BAGEL among the five, with 0.366 compared with 0.942 on SenseNova-U1, and it remains the weakest after training. In summary, Prism preserves most OneIG capabilities, improves the dimensions aligned with its training objectives, and causes small decreases on Alignment and Text, which lie outside these objectives.

\begin{table*}[h]
\caption{TIIF-Bench and editing results. TIIF-Bench uses a short and a long prompt variant per prompt with Qwen2.5-VL-72B as the judge. ImgEdit-Bench scores 1--5 and GEdit-Bench (English) uses VIEScore on a 10-point scale. The better value is in bold.}
\label{tab:tiif_edit}
\centering
\setlength{\tabcolsep}{8pt}
\begin{tabular}{c c cccccc}
\multirow{2}{*}{\textbf{Model}} & \multirow{2}{*}{\textbf{Prism}} & \multicolumn{2}{c}{\textbf{TIIF-Bench}} & \multicolumn{3}{c}{\textbf{Editing}} \\
\cmidrule(lr){3-4} \cmidrule(lr){5-7}
& & \textbf{Short} & \textbf{Long} & \textbf{GEdit-S} & \textbf{GEdit-Q} & \textbf{ImgEdit} \\
\midrule

\multirow{2}{*}{SenseNova-U1}
& \cellr{\xmark}
& \cellg{0.807}{}
& \cellg{\textbf{0.805}}{}
& \cellg{\textbf{7.37}}{}
& \cellg{\textbf{6.79}}{}
& \cellg{\textbf{3.58}}{} \\
{}
& \cellr{\cmark}
& \cellg{\textbf{0.812}}{}
& \cellg{0.794}{}
& \cellg{7.23}{}
& \cellg{6.52}{}
& \cellg{3.47}{} \\
\midrule

\multirow{2}{*}{\cellm{BAGEL}}
& \cellr{\xmark}
& \cellg{\textbf{0.699}}{}
& \cellg{\textbf{0.699}}{}
& \cellg{\textbf{6.89}}{}
& \cellg{\textbf{6.93}}{}
& \cellg{3.33}{} \\
{}
& \cellr{\cmark}
& \cellg{0.664}{}
& \cellg{0.681}{}
& \cellg{6.82}{}
& \cellg{6.87}{}
& \cellg{\textbf{3.38}}{} \\
\end{tabular}
\end{table*}

\noindent\textbf{TIIF-Bench.} TIIF-Bench evaluates instruction following under short and long prompt variants, scored by Qwen2.5-VL-72B over eleven sub-attributes \citep{tiif}. As shown in Table~\ref{tab:tiif_edit}, on SenseNova-U1 Prism slightly increases the short score and slightly decreases the long score, while on BAGEL both scores decrease. On SenseNova-U1, the basic sub-attributes improve under both variants, the advanced ones improve under short prompts and decrease under long ones, and style control decreases under both.

These patterns follow from how the training organizes generation. Every image in a trajectory is generated with the preceding text and images as context and realizes only the incremental state of the current step, so the training reinforces continuation from rich context. A short prompt is close to realizing a single simple state and benefits. A long prompt instead requires satisfying all constraints in one generation from text alone, and style control requires the same one-shot fidelity, so neither is reinforced and both decrease. On BAGEL, which starts weaker on single-image instruction following, the gains do not materialize and the decreases spread across nearly all sub-attributes.

\noindent\textbf{Editing.} ImgEdit-Bench and GEdit-Bench evaluate instruction-based editing, scored by ImgEdit-Judge and VIEScore respectively \citep{imgedit,gedit}. As shown in Table~\ref{tab:tiif_edit}, editing results are mixed. On SenseNova-U1 both benchmarks decrease slightly, with the largest per-task decrease on style change. On BAGEL, ImgEdit-Bench slightly increases, and the increases concentrate on content-oriented operations such as extraction and person editing, while GEdit-Bench changes by less than 0.07.

The transfer is limited because the two tasks use the conditioning image differently. In a trajectory, the preceding images serve as state history, and each new image re-renders the updated state, inheriting entities and progress semantically while sharing few pixels with its predecessors. Instruction-based editing instead keeps most pixels of the input image and changes only the specified region. The training therefore reinforces semantic continuation conditioned on images but never pixel-level preservation, which explains why content-oriented operations improve while tasks that demand pixel fidelity stay flat or decrease. Obtaining editing gains requires editing-specific supervision as in editing-oriented methods such as Uni-Edit \citep{uniedit}.

\section{Detailed Reliability Judgment and Routing Analysis}
\label{app:detailed reliability}

We further examine the reliability judgments and training targets underlying the results in Table~\ref{tab:overall_benchmarks}. All variants start from 2,395 prompt seeds and use the same one-epoch optimization budget and learning rate. Base w/ Self-Training optimizes unmodified self-generated rollouts with $\mathcal{L}_{\mathrm{interleave}}$. Prism w/ Random Routing preserves the number of trajectories assigned to each action within a minibatch but permutes their assignments independently of the judged reliability states. Prism w/o Rewriting retains reliability-based routing but trains trajectories assigned to t2i and i2t on their unmodified rollouts instead of constructing rewritten targets.

Full Prism outperforms Base w/ Self-Training and Prism w/ Random Routing on all seven metrics in Table~\ref{tab:overall_benchmarks}, and Prism w/o Rewriting on six of seven. Relative to Random Routing, it improves GenEval from 87.6 to 89.2, MME from 2297.5 to 2311.6, and MMVP from 59.3 to 62.0. These comparisons support the value of matching training actions to judged reliability states and constructing rewritten targets, beyond direct self-training or preserving the overall action mixture alone. We next examine the frequency of each reliability state, the evolution of routing during training, and the verification of rewritten targets. Tables~\ref{tab:action_distribution} and \ref{tab:verification_funnel} describe the first rollout of each prompt seed, whereas Table~\ref{tab:routing_dynamics} describes all rollouts judged during the training epoch.

\subsection{Empirical Routing Distribution}

Table~\ref{tab:action_distribution} reports the routing decisions for the first rollout of each of the 2,395 SenseNova-U1 prompt seeds. Both chains pass in 52.2\% of these rollouts, while both fail in 14.7\%. The remaining 794 rollouts (33.2\%) have one reliable and one unreliable chain: 571 (23.8\%) are assigned to t2i and 223 (9.3\%) to i2t. Thus, asymmetric reliability occurs in both directions. Trajectory-specific routing allows Prism to condition each correction on the chain judged reliable, rather than applying a fixed rewriting direction to all asymmetric cases.

\begin{table}[h]
\centering
\caption{Routing decisions for the first rollout of each of the 2,395 SenseNova-U1 prompt seeds.}
\label{tab:action_distribution}
\small
\setlength{\tabcolsep}{8pt}
\begin{tabular}{llrr}
Action & Reliability state & Count & Share \\
\midrule
Pass & both chains reliable & 1,250 & 52.2\% \\
t2i & text reliable, image flawed & 571 & 23.8\% \\
i2t & text flawed, image reliable & 223 & 9.3\% \\
Fail & both chains flawed & 351 & 14.7\% \\
\end{tabular}
\end{table}

\subsection{Routing Dynamics during Training}

Table~\ref{tab:routing_dynamics} tracks routing across the 38 updates of the one-epoch Full Prism run. Unlike Table~\ref{tab:action_distribution}, it includes all rollouts judged during training. When a rewritten candidate fails verification, that candidate is discarded and a replacement rollout is generated from a new prompt. The replacement is judged anew and may be assigned to any of the four actions. Consequently, the stage-wise percentages characterize the online training stream rather than the fixed first-rollout cohort. The judge remains a frozen copy of the base model throughout.

The routing distribution changes over the epoch. From S1 to S5, the Pass share rises from 49.3\% to 59.3\%, while the Fail share falls from 17.9\% to 9.7\%. The i2t share declines from 11.5\% to 6.8\%; in contrast, t2i rises initially and remains near one quarter of judged rollouts in later stages. The text-chain pass rate increases by 12.9 points, from 70.6\% to 83.5\%, whereas the image-chain pass rate increases by 5.3 points, from 60.8\% to 66.1\%. With the judge held fixed, these trends are consistent with improving rollout reliability, particularly on the text side, while image-side failures remain a substantial source of correction targets. Because failed candidates are replaced with newly sampled prompts, the stage-wise changes also reflect the composition of the online rollout stream.

\begin{table*}[h]
\centering
\caption{Routing dynamics across the 38 updates of one epoch on SenseNova-U1. Statistics include all judged rollouts, including replacements generated after failed rewrite verification. The judge is frozen throughout. Text and image pass rates are computed from the corresponding routing categories. All values are percentages.}
\label{tab:routing_dynamics}
\small
\setlength{\tabcolsep}{5pt}
\begin{tabular}{llcccccc}
\multirow{2}{*}{Stage} & \multirow{2}{*}{Updates} & \multicolumn{4}{c}{Routing Distribution} & \multicolumn{2}{c}{Judge Pass Rate} \\
\cmidrule(lr){3-6} \cmidrule(lr){7-8}
& & Pass & t2i & i2t & Fail & Text & Image \\
\midrule
S1 & 0--7   & 49.3\% & 21.3\% & 11.5\% & 17.9\% & 70.6\% & 60.8\% \\
S2 & 8--15  & 51.6\% & 25.1\% & 8.7\%  & 14.6\% & 76.7\% & 60.3\% \\
S3 & 16--23 & 55.7\% & 25.2\% & 7.7\%  & 11.4\% & 80.9\% & 63.4\% \\
S4 & 24--30 & 57.6\% & 24.7\% & 7.1\%  & 10.6\% & 82.3\% & 64.7\% \\
S5 & 31--37 & 59.3\% & 24.2\% & 6.8\%  & 9.7\%  & 83.5\% & 66.1\% \\
\end{tabular}
\end{table*}

\subsection{Verification of Rewritten Targets}

A trajectory assigned to t2i or i2t is used as a positive rewritten target only if its correction passes verification. Table~\ref{tab:verification_funnel} reports verification outcomes for the rewriting candidates arising from the first-rollout cohort in Table~\ref{tab:action_distribution}. Of the 571 t2i candidates, 376 (65.8\%) are accepted and 195 (34.2\%) are discarded. Of the 223 i2t candidates, 117 (52.5\%) are accepted and 106 (47.5\%) are discarded. The lower acceptance rate for i2t shows that the two directions have different verification outcomes; neither set of generated corrections is admitted wholesale. Discarded candidates do not become positive training targets and are replaced with newly generated rollouts, which are judged and routed independently.

Verification also entails different amounts of review in the two directions. The first-rollout t2i candidates require 1,474 $I^{+}$ review calls, compared with 320 $T^{+}$ review calls for i2t candidates. These counts include retries, rather than additional training trajectories. Together, the acceptance rates and review counts characterize verification as a target-selection step: routing identifies a chain to correct, while verification determines whether the resulting correction can be used for training.

\begin{table}[t]
\centering
\caption{Verification outcomes for rewriting candidates arising from the first-rollout cohort in Table~\ref{tab:action_distribution}. Accepted and discarded counts are candidate-level terminal outcomes; verification calls include retries for these candidates.}
\label{tab:verification_funnel}
\small
\setlength{\tabcolsep}{6pt}
\begin{tabular}{lcccc}
Rewrite & Candidates & Accepted & Discarded & Verification Calls \\
\midrule
t2i & 571 & 376 (65.8\%) & 195 (34.2\%) & 1,474 ($I^{+}$ reviews) \\
i2t & 223 & 117 (52.5\%) & 106 (47.5\%) & 320 ($T^{+}$ reviews) \\
\end{tabular}
\end{table}

\subsection{Judge Reliability, Contamination, and External-Judge Comparison}
\label{app:judge_reliability}

Routing and verification depend on judgments that may themselves be incorrect. We therefore compare the frozen SenseNova-U1 judge used by Full Prism with Qwen3.7 Plus in a separate judge comparison. We sample 200 training rollouts and obtain human annotations for their complete text and image chains, yielding 400 chain-level judgments. A positive judge prediction denotes a chain judged reliable. Table~\ref{tab:judge_confusion} reports agreement with the human annotations and the false discovery rate (FDR), defined as $\mathrm{FP}/(\mathrm{TP}+\mathrm{FP})$. FDR measures the fraction of chains accepted as reliable that are flawed according to the annotations, and is therefore directly relevant to the quality of accepted teacher signals.

\begin{table*}[h]
\centering
\caption{Chain-level agreement with human annotations on 200 rollout trajectories. Each text or image row contains 200 judgments; the \textit{All} rows aggregate 400 judgments. FDR is $\mathrm{FP}/(\mathrm{TP}+\mathrm{FP})$.}
\label{tab:judge_confusion}
\small
\setlength{\tabcolsep}{8pt}
\begin{tabular}{llrrrrrrr}
\multirow{2}{*}{Judge}
& \multirow{2}{*}{Chain}
& \multicolumn{4}{c}{Human Agreement}
& \multicolumn{3}{c}{Judge Quality} \\
\cmidrule(lr){3-6} \cmidrule(lr){7-9}
& & TP & FP & TN & FN & Acc. & F1 & FDR \\
\midrule
\multirow{3}{*}{SenseNova-U1 (frozen)}
& Text & 113 & 18 & 47 & 22 & 80.0\% & 85.0\% & 13.7\% \\
& Image & 95 & 17 & 64 & 24 & 79.5\% & 82.3\% & 15.2\% \\
& \textit{All} & 208 & 35 & 111 & 46 & 79.8\% & 83.7\% & 14.4\% \\
\addlinespace[3pt]
\multirow{3}{*}{Qwen3.7 Plus}
& Text & 128 & 24 & 41 & 7 & 84.5\% & 89.2\% & 15.8\% \\
& Image & 87 & 7 & 74 & 32 & 80.5\% & 81.7\% & 7.4\% \\
& \textit{All} & 215 & 31 & 115 & 39 & 82.5\% & 86.0\% & 12.6\% \\
\end{tabular}
\end{table*}

Qwen3.7 Plus has higher pooled accuracy and F1 than the frozen judge: 82.5\% versus 79.8\% and 86.0\% versus 83.7\%, respectively. However, its advantage does not extend uniformly to the chains accepted as reliable. The frozen judge has a lower text-chain FDR (13.7\% versus 15.8\%), while Qwen3.7 Plus has a lower image-chain FDR (7.4\% versus 15.2\%). Thus, higher aggregate agreement with human annotations does not imply lower contamination among accepted signals in both modalities.

The consequences of judgment errors also depend on their type. A false negative may reject a usable teacher chain; if the other chain is accepted, it can additionally trigger an unnecessary rewrite. When both chains are rejected, an otherwise reliable text chain may receive the unlikelihood objective. Fail accounts for 14.7\% of the first rollouts in Table~\ref{tab:action_distribution}, although this share does not measure the frequency of such erroneous assignments. A false positive can instead admit a flawed chain as a teacher signal. For rewritten targets, verification provides an additional gate: among the first-rollout candidates in Table~\ref{tab:verification_funnel}, it discards 34.2\% of t2i and 47.5\% of i2t candidates. These measurements establish neither error-free routing nor error-free accepted targets, but identify where judgment noise enters the training pipeline and where verification filters its outputs.

The i2t results connect this analysis to the loss ablation. Replacing $\mathcal{L}_{\mathrm{i2t}}$ with $\mathcal{L}_{\mathrm{interleave}}$ on the corresponding unmodified rollouts lowers MMVP from 62.0 to 56.7 (Table~\ref{tab:ablation}). Meanwhile, verification accepts 117 of the 223 first-rollout i2t candidates and rejects 106 (Table~\ref{tab:verification_funnel}). The ablation supports the contribution of the i2t training objective relative to direct training on the same rollouts, while the verification results show that its rewritten targets are filtered before use.

\begin{figure*}[h]
\centering
\includegraphics[width=0.88\textwidth]{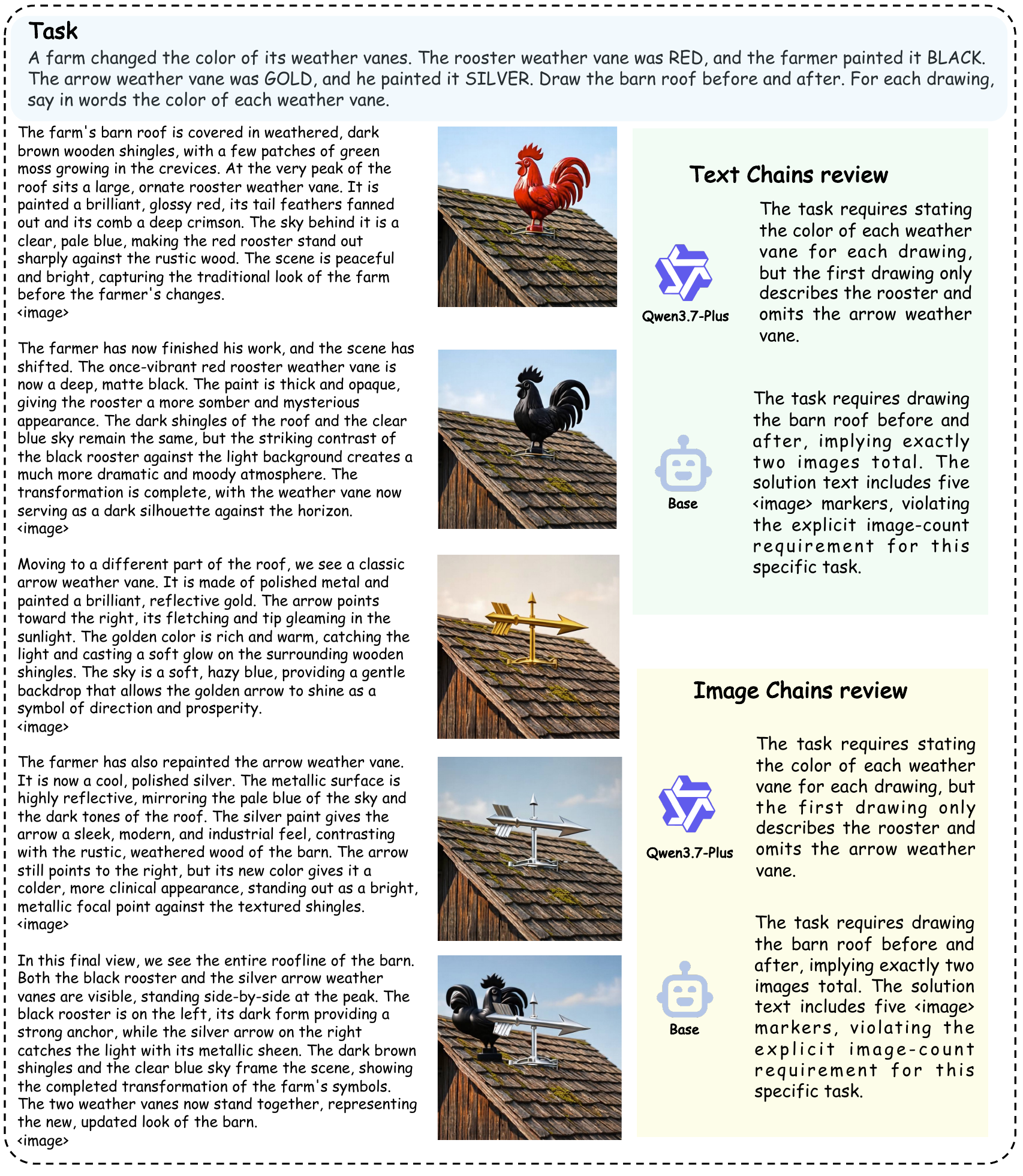}
\caption{Different judgment emphases of Qwen3.7 Plus and SenseNova-U1. Qwen3.7 Plus identifies a local omission, while SenseNova-U1 identifies an image-count violation in the complete trajectory.}
\label{fig:judge_case}
\end{figure*}

We also compare the downstream results obtained with the two judges. As shown in Table~\ref{tab:judge_downstream}, the external judge yields slightly higher MMBench and MMMU scores, whereas the frozen SenseNova-U1 judge yields higher MME, POPE F1, MMVP, GenEval, and DPG; the two tie on MMStar. Figure~\ref{fig:judge_case} illustrates a corresponding difference in judgment emphasis: Qwen3.7 Plus identifies a local omission in the text chain, while SenseNova-U1 identifies an image-count violation in the complete trajectory. The external judge therefore changes the resulting capability profile rather than improving it uniformly. Together with the comparison against Random Routing in Table~\ref{tab:overall_benchmarks}, these results support the use of a frozen internal judge as an informative, albeit imperfect, routing signal.

\begin{table*}[h]
\centering
\caption{Effects of internal and external judges on SenseNova-U1. POPE reports F1.}
\label{tab:judge_downstream}
\small
\setlength{\tabcolsep}{5pt}
\begin{tabular}{lcccccccc}
Judge & MME & MMBench & MMMU MC & MMStar & POPE & MMVP & GenEval & DPG \\
\midrule
Base
& 2282.6 & 86.1 & 59.0 & 65.1 & 84.0 & 58.0 & 85.8 & 86.1 \\
SenseNova-U1
& \textbf{2311.6} & 87.3 & 60.6 & \textbf{66.5} & \textbf{86.6} & \textbf{62.0} & \textbf{89.2} & \textbf{87.3} \\
Qwen3.7 Plus
& 2304.1 & \textbf{87.4} & \textbf{60.7} & \textbf{66.5} & 86.3 & 59.3 & 88.9 & 86.9 \\
\end{tabular}
\end{table*}

\section{The Impact of Freezing Strategy}
\label{app:freezing_strategy}

We examine the role of freezing the judge during online training by comparing Full Prism with a variant in which the judge is updated jointly with the rollout model. The comparison uses the same training setup and tracks two quantities. First, the routing distribution records how the judge assigns generated trajectories to the four actions; the text- and image-chain pass rates are computed from these assignments. Second, the zero-image rate is the proportion of generated interleaved trajectories containing no image output and serves as a measure of image-generation collapse. The jointly updated run terminates after 34 updates, whereas the frozen-judge run completes all 38 updates.

Table~\ref{tab:freezing_dynamics} shows that the two settings have similar routing distributions in the first stage but subsequently diverge. With the frozen judge, the text-chain pass rate increases from 70.6\% to 83.5\% and the image-chain pass rate from 60.8\% to 66.1\%, while all four actions remain active throughout the epoch. The Frozen judge rows reproduce the online training statistics in Table~\ref{tab:routing_dynamics}. With joint updates, the Pass share rises from 50.6\% in the first stage to 87.4\% in updates 16--23, while i2t and Fail fall to 4.8\% and 1.0\%. The increasing concentration in Pass indicates that joint updates substantially change the routing signal.

This concentration does not persist as stable generation. In the final two updates of the jointly updated run, the Pass share falls to 46.0\% and t2i rises to 54.0\%, alongside a drop in the image-chain pass rate to 46.0\%. Table~\ref{tab:zero_image_rate} shows that its final zero-image rate reaches 70.59\%, compared with 0.00\% for the frozen-judge run. Thus, the high intermediate pass rates of the jointly updated judge do not correspond to sustained image-generation quality. In this comparison, freezing the judge maintains a stable routing reference and avoids the observed collapse.

\begin{table*}[h]
\centering
\caption{Routing distributions and judge pass rates across training stages under frozen and jointly updated judges. Text and image pass rates are computed from the corresponding routing categories. All values are percentages. The jointly updated run terminates after 34 updates.}
\label{tab:freezing_dynamics}
\small
\setlength{\tabcolsep}{5pt}
\begin{tabular}{llcccccc}
\multirow{2}{*}{Judge Setting} & \multirow{2}{*}{Updates}
& \multicolumn{4}{c}{Routing Distribution}
& \multicolumn{2}{c}{Judge Pass Rate} \\
\cmidrule(lr){3-6} \cmidrule(lr){7-8}
& & Pass & t2i & i2t & Fail & Text & Image \\
\midrule
\multirow{5}{*}{Frozen judge (Ours)}
& 0--7   & 49.3 & 21.3 & 11.5 & 17.9 & 70.6 & 60.8 \\
& 8--15  & 51.6 & 25.1 & 8.7  & 14.6 & 76.7 & 60.3 \\
& 16--23 & 55.7 & 25.2 & 7.7  & 11.4 & 80.9 & 63.4 \\
& 24--30 & 57.6 & 24.7 & 7.1  & 10.6 & 82.3 & 64.7 \\
& 31--37 & 59.3 & 24.2 & 6.8  & 9.7  & 83.5 & 66.1 \\
\midrule
\multirow{5}{*}{Jointly updated judge}
& 0--7   & 50.6 & 20.8 & 13.2 & 15.4 & 71.4  & 63.8 \\
& 8--15  & 76.0 & 14.4 & 4.8  & 4.8  & 90.4  & 80.8 \\
& 16--23 & 87.4 & 6.9  & 4.8  & 1.0  & 94.3  & 92.2 \\
& 24--31 & 82.2 & 17.1 & 0.5  & 0.2  & 99.3  & 82.7 \\
& 32--33 & 46.0 & 54.0 & 0.0  & 0.0  & 100.0 & 46.0 \\
\end{tabular}
\end{table*}

\begin{table}[h]
\centering
\caption{Effect of freezing the judge on image-generation stability during online training. Lower zero-image rates are better.}
\label{tab:zero_image_rate}
\small
\setlength{\tabcolsep}{8pt}
\begin{tabular}{lcc}
Judge Setting & Initial Zero-Image Rate & Final Zero-Image Rate \\
\midrule
Frozen judge (Ours) & 0.00\% & 0.00\% \\
Jointly updated judge & 0.00\% & 70.59\% \\
\end{tabular}
\end{table}

\section{Implementation of Training Objectives}

Equation~3 defines the objective for one interleaved trajectory $\tau=(T_1,I_1,\ldots,T_K,I_K)$, where $P$ is the task prompt, $T_i$ is the $i$-th text segment, and $I_i$ is the image generated after $T_i$. Each step follows a text-first, then image order. The trajectory-level objective sums a text loss and an image loss over the $K$ interleaved steps.

For the text side, let $T_i=(t_{i,1},\ldots,t_{i,|T_i|})$ denote the token sequence of the $i$-th text segment. We use standard token-level next-token cross-entropy:
\begin{equation}
\mathcal{L}^{(i)}_{\mathrm{und}}
=
-\sum_{n=1}^{|T_i|}
\log p_{\theta}(t_{i,n}\mid P,T_{<i},I_{<i},t_{i,<n}),
\end{equation}
where $T_{<i}=(T_1,\ldots,T_{i-1})$ is the sequence of preceding text segments, $I_{<i}=(I_1,\ldots,I_{i-1})$ is the sequence of preceding images, and $t_{i,<n}=(t_{i,1},\ldots,t_{i,n-1})$ are the previous tokens in the current text segment.

For the image side, let $z_{i,1}$ denote the clean target visual representation of image $I_i$ under the native image representation of the backbone. This may be a latent code, patch representation, or another backbone-specific continuous representation. Let $z_{i,0}\sim p_0$ denote a noise sample of the same shape drawn from a fixed noise distribution. We sample an interpolation time $s\sim\mathcal{U}[0,1]$ and define the linear flow-matching path
\begin{equation}
z_{i,s}=(1-s)z_{i,0}+sz_{i,1},
\end{equation}
with target velocity
\begin{equation}
v_i^{\star}=z_{i,1}-z_{i,0}.
\end{equation}

Let
\begin{equation}
c_i=(P,T_{\leq i},I_{<i}),
\qquad
T_{\leq i}=(T_1,\ldots,T_i),
\end{equation}
denote the conditioning context for generating $I_i$. The model predicts a velocity field $v_{\theta}(z_{i,s},s,c_i)$, and the image-generation loss is
\begin{equation}
\mathcal{L}^{(i)}_{\mathrm{gen}}
=
\mathbb{E}_{s\sim\mathcal{U}[0,1],z_{i,0}\sim p_0}
\left[
\left\|
v_{\theta}(z_{i,s},s,c_i)-v_i^{\star}
\right\|_2^2
\right].
\end{equation}

For the both-pass case, both modalities are judged reliable. We keep the original image chain $I=(I_1,\ldots,I_K)$ and reconstruct the full concatenated text chain $T=(t_1,\ldots,t_{|T|})$, where $T$ is obtained by concatenating all text segments $T_1,\ldots,T_K$. The recovery instruction $P_{\mathrm{rec}}$ and image chain jointly condition the reconstruction:
\begin{equation}
\mathcal{L}_{\mathrm{pass}}
=
-\sum_{n=1}^{|T|}
\log p_{\theta}(t_n\mid P_{\mathrm{rec}},I,t_{<n}).
\end{equation}

For the t2i and i2t cases, one modality is corrected and the resulting rewritten trajectory is verified before training. We then apply Equation~3 to the verified corrected trajectory. Thus, every text token is trained with the text loss and every image representation is trained with the image loss.

For the both-fail case, the original trajectory $\tau$ is treated as unreliable and is not used as a positive target. Instead, we apply token-level unlikelihood training to the original text chain:
\begin{equation}
\mathcal{L}_{\mathrm{fail}}(\tau)
=
-\sum_{i=1}^{K}\sum_{n=1}^{|T_i|}
\log\left(1-p_{\theta}(t_{i,n}\mid P,T_{<i},I_{<i},t_{i,<n})\right).
\end{equation}

In implementation, the argument of the logarithm is numerically clipped away from zero for stability. Each trajectory $\tau$ is assigned to exactly one reliability case, pass, t2i, i2t, or fail. During minibatch training, only the objective associated with the assigned case contributes to the loss for that trajectory, while the remaining case-specific objectives are masked out. Eq.~\ref{eq:total} denotes the resulting aggregate minibatch objective.

\section{Verification of Rewritten Trajectories}
\label{rewritten}
For the t2i and i2t cases, Prism first generates a corrected chain from the preserved reliable chain and the detected error. Text-to-image rewriting generates $I^+$ from $P$, $T$, and $r_{\mathrm{img}}$, while image-to-text rewriting generates $T^+$ from $P$, $I$, and $r_{\mathrm{txt}}$. Each rewritten candidate is verified before it is used as a positive training target. The verifier is a frozen copy of the base model. A candidate is regenerated for at most four attempts, including the initial generation and three retries. Candidates that fail all attempts are discarded.

\noindent\textbf{Image-side verification.}
For a rewritten image chain $I^+$ generated from the preserved text chain $T$, the verifier receives the task prompt $P$ and the candidate image sequence. A candidate is accepted only if it contains the same number of images as the original trajectory, satisfies the task requirements for objects, counts, attributes, spatial relations, state changes, and step order, and is judged valid by the verifier.

\begin{verbatim}
You are a strict reviewer auditing only the image sequence
of an interleaved solution and its visual state progression.

Step text is not shown. You see {n_images} images in order.

TASK:
{prompt}

Decide whether this image chain delivers the progression
required by the task.

1. Base each decision on visible evidence.
2. To return invalid, cite one concrete image number and its visible error.
3. The first confirmed visible error determines invalid.
4. Determine from the task whether image count is strict or flexible.
5. Check required objects, counts, attributes, spatial relations,
   actions, visual states, lighting, and state progression.
6. A preview or final state is valid only when it does not conflict
   with an explicitly required stage.
7. Do not reject an image chain solely for style drift, texture noise,
   lighting shifts, background changes, or minor visual variation.

Return invalid for a confirmed error such as no task progress,
wrong image or stage count, wrong quantity, failed state change,
state mismatch, corrupted key subject, premature final state,
state regression, identity swap, internally impossible content,
or failure to convey a required step.

Return exactly one JSON object with a Boolean field named "valid".
\end{verbatim}

\noindent\textbf{Text-side verification.}
For a rewritten text chain $T^+$ generated from the preserved image chain $I$, the verifier receives the task prompt and candidate text. A candidate is accepted only if the verifier returns valid, its interleaved marker structure is valid, its final image reference is valid, and its length is at least half that of the original rollout text.

\begin{verbatim}
You are auditing only the visible text of a no-think interleaved
text-and-image response. Images are audited separately.

TASK:
{prompt}

SOLUTION TEXT:
{text_chain}

Judge whether the written content and image marker structure
satisfy the task.

1. Enforce image-marker count only when the task explicitly specifies it.
2. Required facts, names, labels, directions, emotions, colors, counts,
   and object identities must be stated clearly.
3. The text must cover required objects, stages, relations, or scene changes.
4. Reject only clear text-side contradictions, skipped main events,
   impossible orderings, or unexplained state regression.
5. For exact-one-image tasks, text after the image marker must not
   introduce another scene or visual plan.
6. For tutorials, processes, stories, and lifecycle tasks, check clear
   ordering errors, missing stages, extra stages violating an explicit
   count, or state regression.

Return exactly one JSON object with a Boolean field named "valid".
When invalid, include a verbatim faulty span in the "quote" field.
\end{verbatim}

\section{Additional Understanding Examples}
\label{understanding examples}
\begin{figure*}[h]
    \centering
    \includegraphics[width=\textwidth]{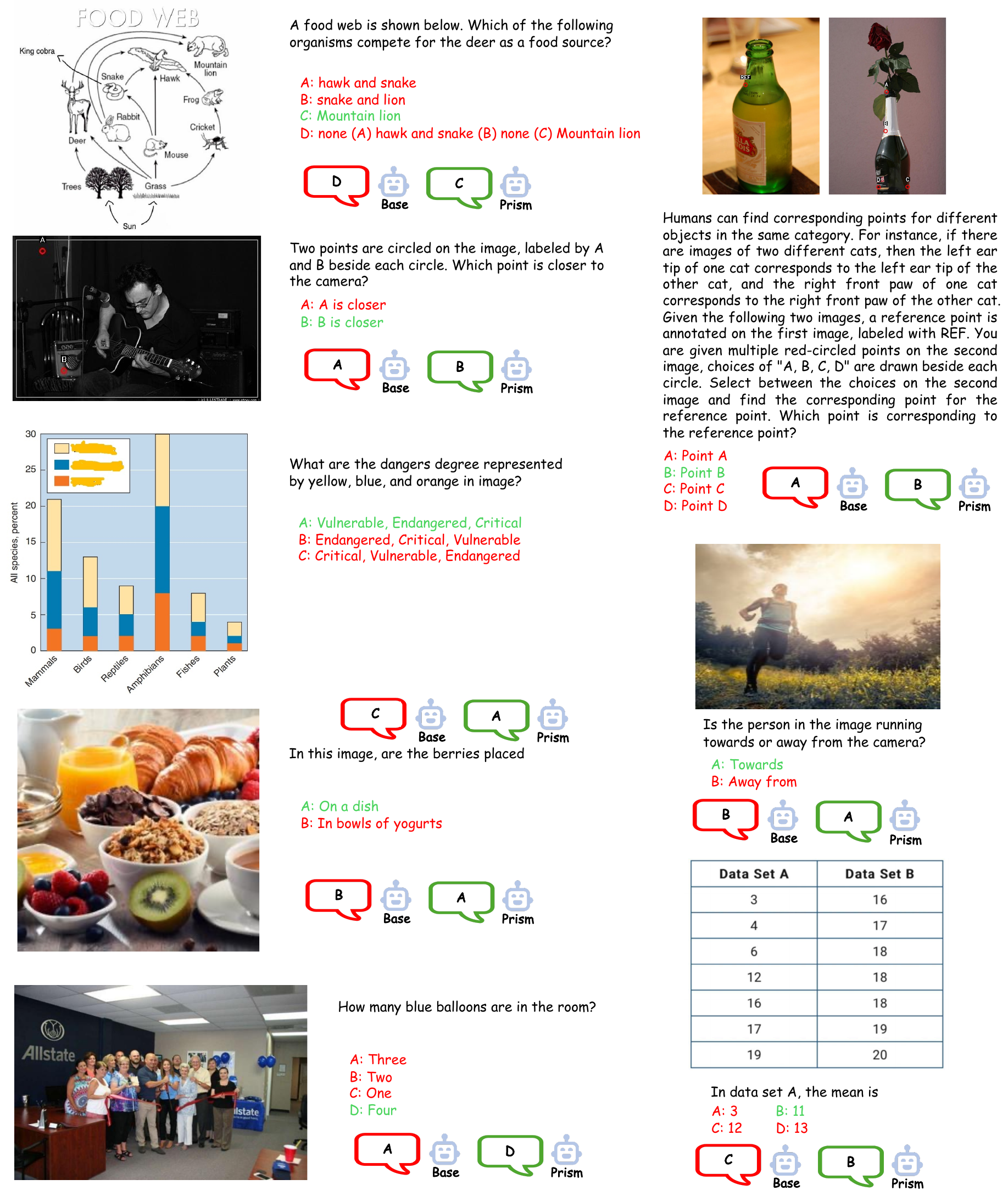}
    \caption{Additional VQA examples. Prism corrects errors made by the base model on questions requiring visual reasoning, chart understanding, correspondence matching, spatial perception, and counting.}
    \label{fig:appendix_vqa}
\end{figure*}

Figure~\ref{fig:appendix_vqa} presents additional visual question answering examples. Compared with the base model, Prism gives correct responses on diverse tasks involving food-web reasoning, depth perception, chart interpretation, semantic correspondence, motion direction, spatial attributes, legend understanding, and object counting. These examples further illustrate the improved visual grounding and multimodal reasoning observed in the quantitative evaluation.

\section{Additional Generation Examples}
\label{genereation examples}
\begin{figure*}[h]
    \centering
    \includegraphics[width=\textwidth]{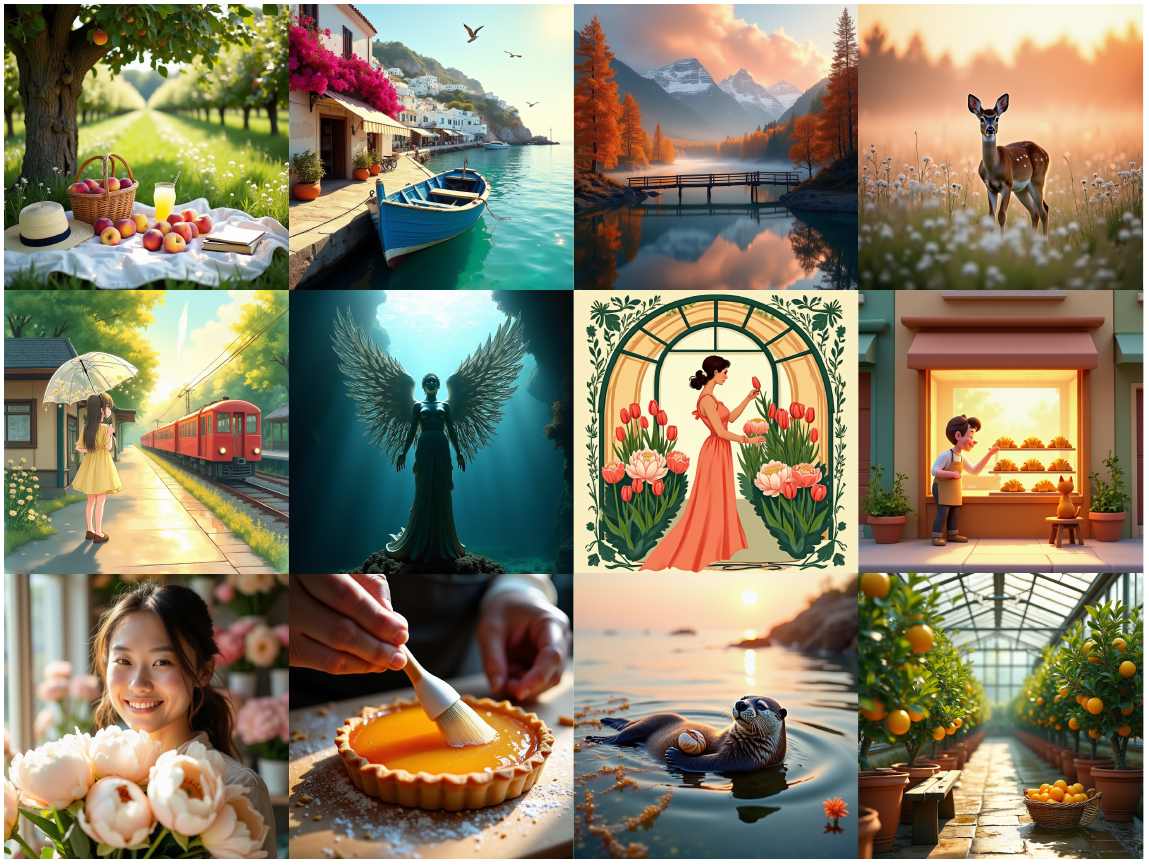}
    \caption{Additional generation examples on BAGEL. The generated images cover diverse subjects, scenes, and visual styles.}
    \label{fig:appendix_bagel_t2i}
\end{figure*}

Figure~\ref{fig:appendix_bagel_t2i} presents additional generation examples on BAGEL. The examples cover diverse visual styles and contents, including realistic photography, landscape scenes, illustration, portraiture, food, wildlife, and indoor environments.

\section{Additional Interleaved Generation Examples}
\label{interleaved examples}
\begin{figure*}[t]
    \centering
    \includegraphics[width=\textwidth]{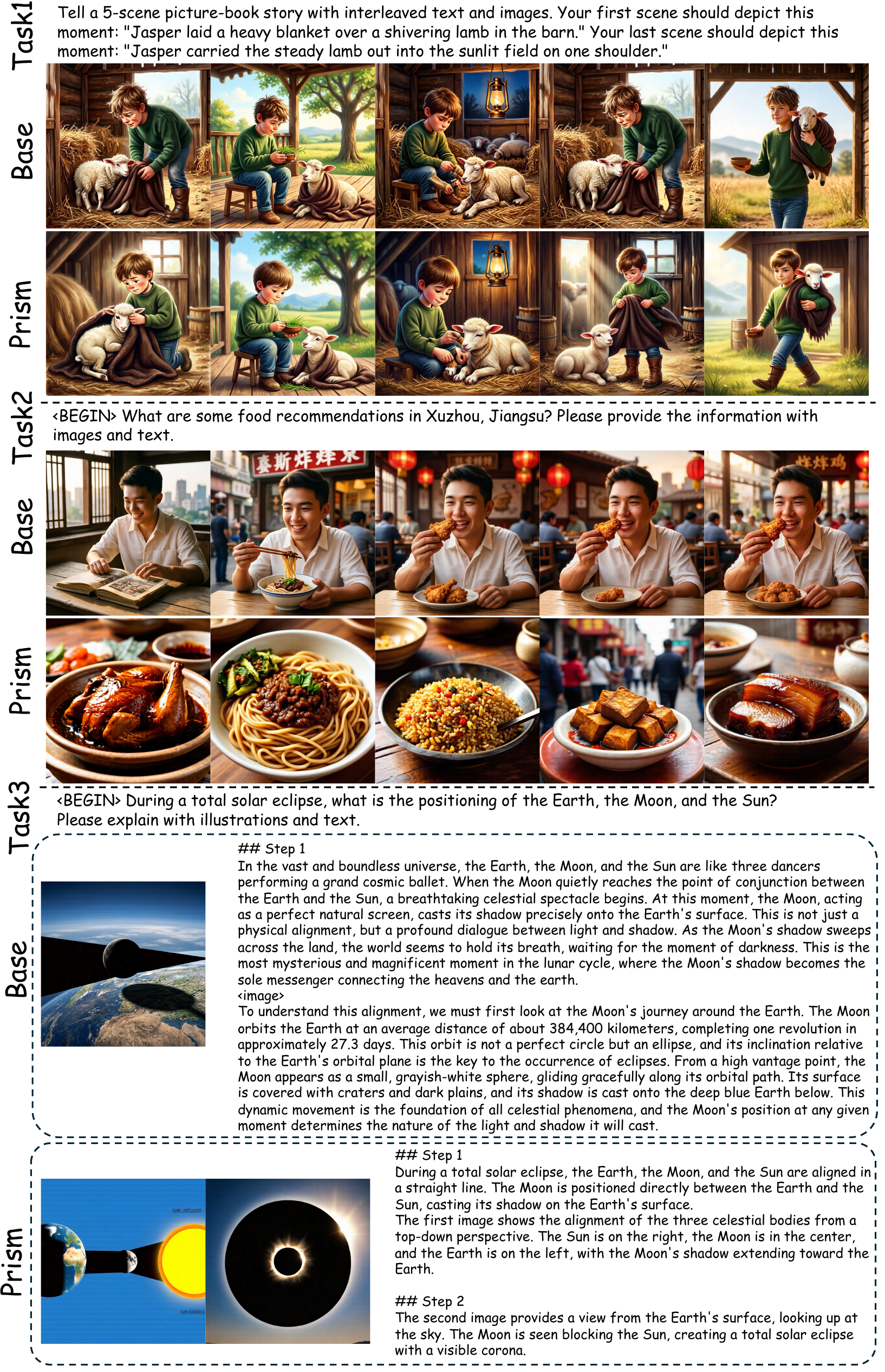}
    \caption{Additional interleaved generation examples. Compared with the base model, Prism produces clearer visual progression in a picture-book story, more diverse images for food recommendations, and visual understanding for a solar eclipse.}
    \label{fig:appendix_interleaved}
\end{figure*}

Figure~\ref{fig:appendix_interleaved} compares interleaved generation by the base model and Prism. In the picture-book task, the base model repeats similar barn scenes, making the visual progression of the story unclear. For food recommendations, it repeatedly generates a person eating while providing limited visual coverage of the requested dishes. In the solar-eclipse example, the base model produces a long text-dominated response with only one illustration, whose cluttered background does not clearly convey the required Earth--Moon--Sun alignment. In contrast, Prism produces distinct visual states for the story, diverse food images for the recommendation, and stepwise text--image explanations of the eclipse.

\end{document}